\def\year{2020}\relax
\documentclass[letterpaper]{article} % DO NOT CHANGE THIS
\usepackage{aaai20}  % DO NOT CHANGE THIS
\usepackage{times}  % DO NOT CHANGE THIS
\usepackage{helvet} % DO NOT CHANGE THIS
\usepackage{courier}  % DO NOT CHANGE THIS
\usepackage[hyphens]{url}  % DO NOT CHANGE THIS
\usepackage{graphicx} % DO NOT CHANGE THIS
\usepackage{graphicx}  % DO NOT CHANGE THIS
\usepackage{xcolor}
\usepackage{amsmath,amssymb,amsfonts,amsthm,mathtools}
\usepackage{bm}
\usepackage{tikz}
\usetikzlibrary{fit,positioning,calc}
\usepackage{subcaption}
\usepackage{booktabs}
\usepackage{multirow}
\usepackage{algorithm}
\usepackage{algorithmic}

\newcommand{\citet}{\cite}
\newcommand{\citep}{\cite}

\newcommand{\argmax}{\operatornamewithlimits{argmax}}

\def\reals{\mathbb{R}}
\def\Y{\mathcal{Y}}
\def\y{\bm y}
\def\X{\bm X}
\def\x{\bm x}
\def\f{\bm f}
\def\h{\bm h}

\def\t{\bm \theta}

\def\w{\bm w}

\def\v{\bm v}

\definecolor{c_yellow}{RGB}{253,237,106}
\definecolor{c_red}{HTML}{FD7F7C}
\definecolor{c_blue}{HTML}{A7DFF8}

\title{Deep Mixed Effect Model using Gaussian Processes: \\	
A Personalized and Reliable Prediction for Healthcare}
\author{
Ingyo Chung,\textsuperscript{\rm 1}
Saehoon Kim,\textsuperscript{\rm 2}
Juho Lee,\textsuperscript{\rm 2}
Kwang Joon Kim,\textsuperscript{\rm 3}
Sung Ju Hwang,\textsuperscript{\rm 1,2}
Eunho Yang\textsuperscript{\rm 1,2}\\
\textsuperscript{\rm 1}KAIST, 
\textsuperscript{\rm 2}AITRICS, 
\textsuperscript{\rm 3}Yonsei University College of Medicine, South Korea\\
\{jik0730,sjhwang82,eunhoy\}@kaist.ac.kr, \{shkim,juho\}@aitrics.com, preppie@yuhs.ac
}
\begin{document}

\maketitle

\begin{abstract}
We present a personalized and reliable prediction model for healthcare, which can provide individually tailored medical services such as diagnosis, disease treatment, and prevention. Our proposed framework targets at making personalized and reliable predictions from time-series data, such as Electronic Health Records (EHR), by modeling two complementary components: i) a shared component that captures global trend across diverse patients and ii) a patient-specific component that models idiosyncratic variability for each patient. To this end, we propose a composite model of a deep  neural network to learn complex global trends from the large number of patients, and Gaussian Processes (GP) to probabilistically model individual time-series given relatively small number of visits per patient. We evaluate our model on diverse and heterogeneous tasks from EHR datasets and show practical advantages over standard time-series deep models such as pure Recurrent Neural Network (RNN).
\end{abstract}

%%%%%%%%%%%%%%%%%%%%%%%%%%%%%%%%%%%%%%%%%%%%%%%%%%%%%%%%%%%%%%%%%%%%

\section{Introduction}\label{sec:introduction}

Precision medicine, which aims to provide \emph{individually} tailored medical services such as diagnosis, disease treatment, and prevention, is an ultimate goal in healthcare. While rendered difficult in the past, nowadays it is becoming increasingly realizable due to the advances in data-driven approaches such as machine learning. Especially, recent widespread use of Electronic Health Record (EHR), a systematic collection of diverse clinical records of patients, has encouraged machine learning researchers to explore various clinical inferences (such as heart failure risk prediction~\citep{retain}, sepsis prediction~\citep{sepsis}, and physiological time-series analysis~\citep{mtgp1}, to name a few) based on the records of personal medical history to improve the quality of clinical cares \citep{rnn1,rnn3,rnn2}. 

\begin{figure}[t]
\centering
\vspace{-.5cm}
\includegraphics[width=0.8\linewidth]{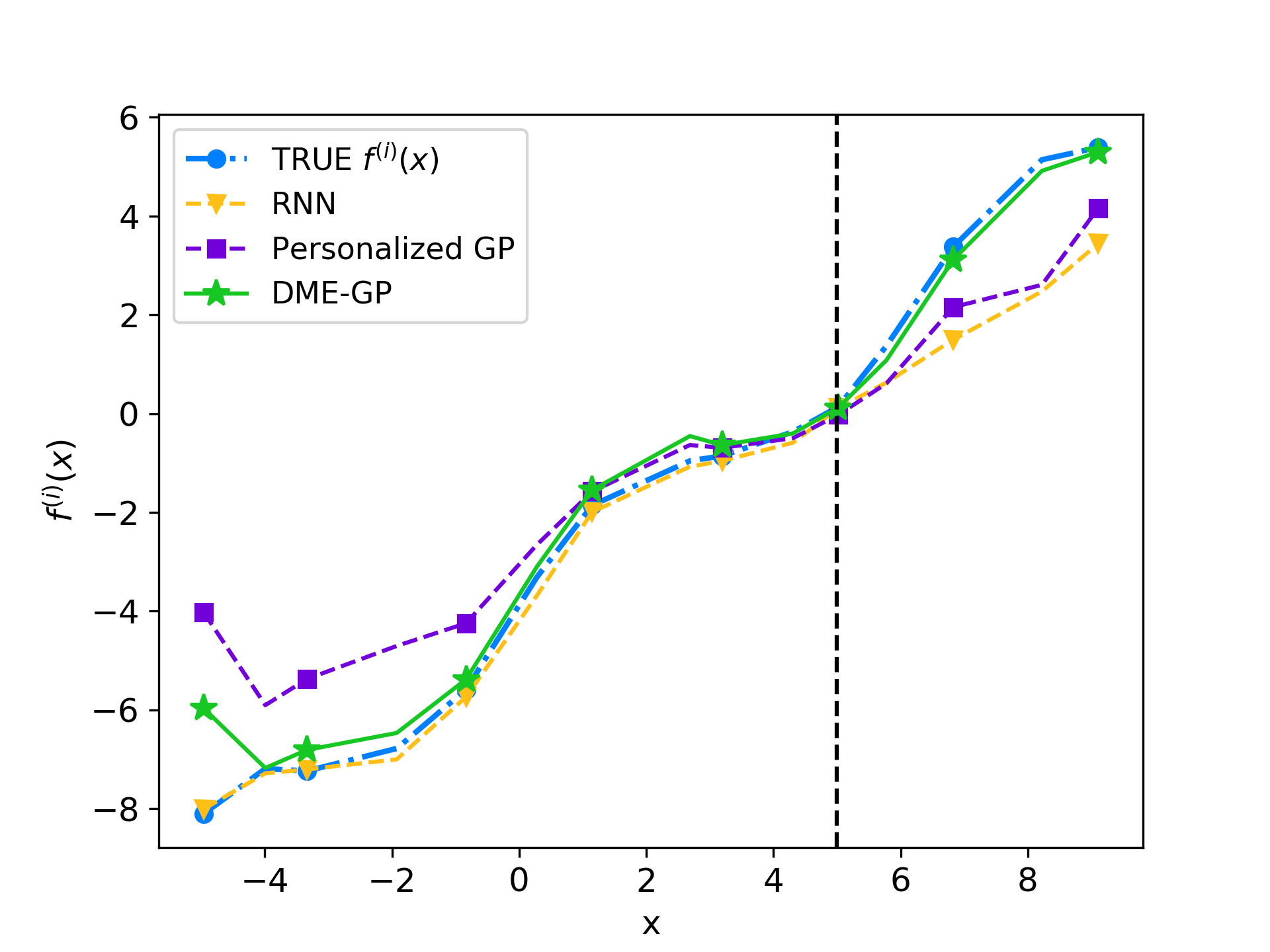}
\vskip -.3cm
\caption{\small \textbf{Motivating experiment.} We generate data for a patient $i$ by $f^{(i)}(x)=x+\sin(x)+\epsilon^{(i)}$ where $\epsilon^{(i)} \sim \mathcal{N} (\mu_i, \sigma_i^2)$ as deviation from others. RNN fails to provide personalized predictions on unseen input range (after dashed line) due to lack of personalization. Personalized GP dismisses the global trend. The behavior of other baselines are shown in Appendix though each has its disadvantage, compared to our model (denoted as DME-GP).}
\label{FigIntro}
\vspace{-.5cm}
\end{figure}

As we have seen the recent huge success of deep learning, one of the most popular choices when working with EHR is to use Recurrent Neural Network (RNN) based models \citep{rnn1,retain}. However, this kind of so-called ``population based'' models (that is, learning a single model for all patients as in RNN) might fail to give \emph{personalized} predictions due to huge variability or heterogeneity among patients originated from diverse (possibly unobserved) sources such as intrinsic differences of patients due to demographical and biological factors, or other environmental factors \citep{hete3,hete1,hete2}. To demonstrate this issue, we illustrate in Figure~\ref{FigIntro} how heterogeneities across patients can impact the overall performances of population based models. In this toy simulation (see Appendix for details), we generate a bunch of patient-specific time-series $f^{(i)}(x)$ ($i$ is patient index) with some globally shared structure and train the model with data from limited region (before dashed line).  Figure~\ref{FigIntro} shows some \emph{test} case. Here we can observe that for the region not used for training (after dashed line), teacher forced RNN tends to make biased predictions to the average of training patients. One possible and the easiest solution towards \emph{personalization} is to model separate functions, one per patient to model the heterogeneity, but in a multitask framework to share the common knowledge across patients. However, deep models typically require large data, making it very challenging to train \emph{separate} models for each patient.

Gaussian Process (GP) is another popular model for time-series as a non-parametric model, hence it might be the proper choice to separately model each patient. Due to its probabilistic nature, GP has additional benefit of representing uncertainty, which is also critical in medical problems for \emph{reliable} prediction. However, when working with large amount of data, modeling exact GP gets computationally challenging since it requires to compute the inverse of covariance matrix across all data points ($O(n^3)$ in exact inference for $n$ data points), although several approximations such as \citet{spgp,spgp2} have been proposed at the expense of performance degradation (We will discuss related multi-task GPs and their computational issues in later sections). Some works such as \citet{pgp1} have proposed to use separate GP for each time-series as an alternative, however these models cannot take advantages from global perspective as simply shown in Figure~\ref{FigIntro}, Personalized GP.

In order to only take benefits from different approaches in multiple time-series, in this paper we propose a family of mixed effect model, Deep Mixed Effect Model using GPs (DME-GP), which combines a deep neural network and GPs. In DME-GP, by leveraging the complementary properties of two ingredients, we use a deep neural network (such as RNN) to capture the global trend from the large number of patients and GP to model idiosyncratic variability of individual patient. Regarding on the choice of the former, the global function should be representationally powerful enough to capture the complex shared trend across large number of patients. Not only that, it should be computationally amenable to handle large number of patients in training as well as inference procedures. Deep models including RNN architecture are the reasonable option to leverage the size of EHR data (in terms of the number of patients). On the other hand for the latter, we deliberately choose GP in order to provide reliable and individualized predictions even with very limited number of data points (i.e., limited number of patient's visits). The use of GP naturally makes our model probabilistic and enables us to obtain the prediction uncertainty, which is another important property for mission-critical clinical tasks.

Our main contribution is threefold:
\vspace{-0.1cm}
\begin{itemize}
\item We propose a frame of mixed effect model for multiple time-series that can leverage the individual benefits of shared global function and local function.  
\vspace{-0.1cm}
\item We present the showcase of our framework for EHR analysis that uses RNN for capturing complex global trend in time-series and GP for personalized and reliable prediction with uncertainty estimate.
\vspace{-0.1cm}
\item We show practical advantages of our model over diverse baselines both on regression (from Physionet Challenge 2012 \citep{physionet}) and classification tasks (from National Health Insurance Service; NHIS), and investigate its reliability, which is essential for mission-critical tasks.
\end{itemize}
\vspace{-0.1cm}

%%%%%%%%%%%%%%%%%%%%%%%%%%%%%%%%%%%%%%%%%%%%%%%%%%%%%%%%%%%%%%%%%%%%

\section{Related Work}\label{sec:related}

Due to space constraint, here we focus to review works modeling EHR. More comprehensive review (e.g. review on combining GPs and deep models) is provided in Appendix.

\vspace{-.3cm}
\paragraph{Multiple Gaussian Processes for EHR}
Gaussian Process models have been actively used in the medical applications thanks to their reliability and versatility. However, using the separate formulation of multiple GPs is preferred due to its computational cost. \citet{pgp1} proposed a multiple GPs formulation to handle missing values caused by sensor artifact or data incompleteness, which is common situation in wearable devices. \citet{pgp2} proposed to use a similar model for diagnosis of Alzheimer's disease, where a population-level GP is adapted to a new patient using domain GPs individually. 

\vspace{-.3cm}
\paragraph{Multi-task Gaussian Process for EHR.}
The previous line of works can be understood as multi-task learning in the sense that the parameters of GPs across patients are shared. However, more systematic way of considering multi-task learning with GPs is to directly learn a shared covariance function over tasks. \citet{mtgp4,mtgp2} proposed Multi-task Gaussian Process (MTGP) that constructs large covariance matrix as a Kronecker product of input and task-specific covariance matrices for multi-task learning. A practical example of applying MTGP in medical situation is given in \citet{mtgp1}. \citet{mtgp3} proposed another approach that shares covariance matrix structured as the linear model of coregionalization (LMC) framework for personalized GPs, which is generalization of \citet{mtgp5}. \citet{sepsis,sepsis2} made use of MTGP for preprocessing of input data fed into RNN. All of this line of works are based on the multi-task GPs that share huge covariance matrix which makes exact inference intractable. There have been some attempts to utilize mean of GP similar to our approach, proposed by \citet{shared-gp1}, \citet{shared-gp2}, and \citet{mean-nn}. However, our model is constructed in distinctive way where we use flexible deep models for shared mean function to capture complex structures, and more importantly, we explicitly construct a single GP for each patient to reflect individual signal.

\vspace{-.3cm}
\paragraph{Deep learning models with EHR.}
Recurrent neural networks (RNN) have recently been gained popularity as means of learning a prediction model on time-series clinical data such as EHR. \citet{rnn1} and \citet{doctorai} proposed to use RNN with Long-Short Term Memory (LSTM) \citep{lstm} and Gated Recurrent Units (GRU) \citep{gru} respectively for multi-label classification of diagnosis codes given multivariate features from EHR. Moreover, the pattern of missingness, which is typical property of EHR, has been exploited in \citet{rnn3} and \citet{rnn2} by introducing missing indicator and the concept of decaying informativeness. \citet{retain} proposed to use RNN for generating attention on which feature and hospital visit the model should attend to, for building an interpretable model, and demonstrated it on heart failure prediction task. While RNN models have shown impressive performance on real-world clinical datasets, deploying them to safety-critical clinical tasks should be done with caution as they lack the notion of confidence, or uncertainty of prediction.

%%%%%%%%%%%%%%%%%%%%%%%%%%%%%%%%%%%%%%%%%%%%%%%%%%%%%%%%%%%%%%%%%%%%

\section{Proposed Method}\label{sec:MTCGP}

\subsection{Problem Formulation}
Suppose dataset $\mathcal{D} := \{ (\X_i, \y_i) \}_{i=1}^P$ consists of $P$ patients and $i$-th patient is represented by a sequence of $T_i$ elements (or visits), that is, $\X_i := [\x_1^{(i)}, \x_2^{(i)}, \hdots, \x_{T_i}^{(i)}]$, and corresponding target values, $\y_i:= [y_1^{(i)}, y_2^{(i)}, \hdots, y_{T_i}^{(i)}]$. The goal of our task in patient modeling is to predict the target value $y_t^{(i)}$ at each time step, given the current input features $\x_t^{(i)}$ and all the previous history: $\{\x_{s}^{(i)}\}_{s=1}^{t-1}$ and $\{y_{s}^{(i)}\}_{s=1}^{t-1}$.

This problem formulation incorporates several problems in EHR analysis such as disease progression modeling (DPM) or learning to diagnose (L2D) \cite{retain}. In DPM, we predict the evolutions of medical codes simultaneously at every time point. Specifically, if we have $r$ different medical codes in our EHR, $\x_t \in \reals^{r}$, which encodes the binary status indicating if each code appears in $t$-visit data, our goal is to predict $\x_{t}$ at every time $t$ given all the previous history $\{\x_s\}_{s=1}^{t-1}$. In L2D, which can be thought of as the special case of DPM, we are interested only in diagnosing of certain disease at the very end of visit sequence.

\subsection{A Framework of Mixed Effect Model}
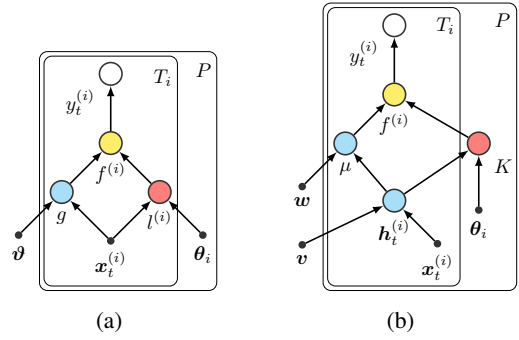
\begin{figure}[t]
\centering
\vspace{-.4cm}
\begin{subfigure}[t]{0.45\linewidth}
	\centering
	\scalebox{0.75}{
	\begin{tikzpicture}
	\tikzstyle{main}=[circle, minimum size = 4mm, thick, draw =black!80, node distance = 8mm]
	\tikzstyle{connect}=[-latex, thick]
	\tikzstyle{box}=[rounded corners, draw=black!100]
	  \node[main, fill=black!80, minimum size=3pt, inner sep=0pt] (vartheta) [label=below:$\bm \vartheta$] { };
	  \node[main] (g) [above right=of vartheta,label=below:$g$, fill=c_blue!100] { };
	  \node[main] (fi) [above right=of g,label=below:$f^{(i)}$, fill=c_yellow!100] { };
	  \node[main] (li) [below right=of fi,label=below:$l^{(i)}$, fill=c_red!100] { };
	  \node[main, fill=black!80, minimum size=3pt, inner sep=0pt] (hit) [below=of $(g)!0.5!(li)$,label=below:$\x^{(i)}_t$] { };
	  \node[main, fill=black!80, minimum size=3pt, inner sep=0pt] (theta) [below right=of li,label=below:$\t_i$] {};
	  \node[main, draw=white!100, opacity=0, text opacity=1, minimum size=3pt, inner sep=0pt, node distance=5mm] (dummy) [below=of hit,label=below:] { };
	  \node[main] (yit) [above=of fi,label=below left:$y^{(i)}_t$] { };
	  \path (vartheta) edge [connect] (g)
	        		(hit) edge [connect] (g)
			(hit) edge [connect] (li)
			(theta) edge [connect] (li)
			(g) edge [connect] (fi)
			(li) edge [connect] (fi)
			(fi) edge [connect] (yit);
	  \node[rounded corners, inner sep=0mm, fit= (g) (hit) (li) (fi) (yit) (dummy),label=above right:$T_i$, xshift=-4.5mm, yshift=3mm] {};
	  \node[rounded corners, inner sep=1mm, y=30cm, draw=black!100, fit= (g) (hit) (li) (fi) (yit) (dummy)] {};
	  \node[rounded corners, inner sep=0mm, fit= (g) (hit) (li) (fi) (yit) (dummy) (theta),label=above right:$P$, xshift=-3.2mm, yshift=1.5mm] {};
	  \node[rounded corners, inner sep=1.8mm, draw=black!100, fit = (g) (hit) (li) (fi) (yit) (dummy) (theta)] {};
	\end{tikzpicture}}
	\caption{}
	\label{PGM1}
\end{subfigure}
\begin{subfigure}[t]{0.45\linewidth}
	\centering
	\scalebox{0.75}{
	\begin{tikzpicture}
	\tikzstyle{main}=[circle, minimum size = 4mm, thick, draw =black!80, node distance = 8mm]
	\tikzstyle{connect}=[-latex, thick]
	\tikzstyle{box}=[rounded corners, draw=black!100]
	  \node[main, fill=black!80, minimum size=3pt, inner sep=0pt] (vartheta) [label=below:$\w$] { };
	  \node[main] (g) [above right=of vartheta,label=below:$\mu$, fill=c_blue!100] { };
	  \node[main] (fi) [above right=of g,label=below:$f^{(i)}$, fill=c_yellow!100] { };
	  \node[main, draw=white!100, opacity=0, text opacity=1] (lid) [below right=of fi,label=below:] { };
	  \node[main, node distance=2mm] (li) [right=of lid,label=below right:$K$, fill=c_red!100] { };
	  \node[main, draw=white!100, opacity=0, text opacity=1, node distance=-1.5mm] (lidd) [right=of li,label=below:] { };
	  \node[main] (hit) [below=of $(g)!0.5!(lid)$,label=below:$\h^{(i)}_t$, fill=c_blue!100] { };
	  \node[main, fill=black!80, minimum size=3pt, inner sep=0pt] (xit) [below right=of hit,label=below:$\x^{(i)}_t$] { };
	  \node[main, fill=black!80, minimum size=3pt, inner sep=0pt, node distance=9mm] (theta) [below=of li,label=below:$\t_i$] {};
	  \node[main, draw=white!100, opacity=0, text opacity=1, minimum size=3pt, inner sep=0pt, node distance=4.5mm] (dummy) [below=of xit,label=below:] { };
	  \node[main, draw=white!100, opacity=0, text opacity=1, minimum size=3pt, inner sep=0pt, node distance=1mm] (dummy2) [right=of xit,label=below:] { };
	  \node[main, fill=black!80, minimum size=3pt, inner sep=0pt, node distance=9mm] (vv) [below=of vartheta,label=below:$\v$] {};
	  \node[main] (yit) [above=of fi,label=below left:$y^{(i)}_t$] { };
	  \path (vartheta) edge [connect] (g)
	        		(hit) edge [connect] (g)
			(hit) edge [connect] (li)
			(xit) edge [connect] (hit)
			(theta) edge [connect] (li)
			(g) edge [connect] (fi)
			(li) edge [connect] (fi)
			(fi) edge [connect] (yit)
			(vv) edge [connect] (hit);
	  \node[rounded corners, inner sep=0mm, fit= (g) (xit) (hit) (fi) (yit) (dummy) (dummy2),label=above right:$T_i$, xshift=-4.5mm, yshift=8mm] {};
	  \node[rounded corners, inner sep=1mm, y=30cm, draw=black!100, fit= (g) (xit) (hit) (fi) (yit) (dummy) (dummy2)] {};
	  \node[rounded corners, inner sep=0mm, fit= (g) (xit) (hit) (li) (fi) (yit) (dummy) (dummy2) (theta) (lidd),label=above right:$P$, xshift=-3.4mm, yshift=4.5mm] {};
	  \node[rounded corners, inner sep=1.8mm, draw=black!100, fit = (g) (xit) (hit) (li) (fi) (yit) (dummy) (dummy2) (theta) (lidd)] {};
	\end{tikzpicture}}
	\caption{}
	\label{PGM2}
\end{subfigure}
\vspace{-.3cm}
\caption{\small (a) A graphical representation of mixed effect framework for EHR analysis in \eqref{Eqn:MEMGeneral}. (b) A graphical representation of DME-GP in \eqref{Eqn:MEMGPRNN}. To emphasize our decomposing framework, we color the global and individual components as blue and red respectively. Final composite model is colored as yellow.}
\vspace{-.5cm}
\end{figure}

Now we provide the general description of our mixed effect framework decomposing the function $f^{(i)}$ for $i$-th patient into two independent functions $g(\cdot)$ and $l^{(i)}(\cdot)$ under the multi-task learning paradigm: 
\begin{align}\label{Eqn:MEMGeneral}
	f^{(i)}(\x_t) = g(\x_t) + l^{(i)}(\x_t)
\end{align}
where we assume $l^{(i)}(\cdot)$ to include random noise. Note that $g(\cdot)$ and $l^{(i)}(\cdot)$ are also called as fixed and random effect respectively in other literature \citep{mixed-effect}. In the framework, $g(\cdot)$ models global trend among the whole diverse patients, and hence it is shared across all patients. On the other hand, $l^{(i)}(\cdot)$ models the patient-specific signal (for $i$-th patient) that is not captured by the global trend $g(\cdot)$. Note that no information is shared across patients through $l^{(i)}(\cdot)$. We highlight that since the framework is generic, both functions can be chosen to be optimal depending on whatever the domain we apply on. Note also that in the traditional multi-task learning, we usually employ this kind of information sharing strategy at the \emph{parameter} level; that is, the parameter vector for each task is represented as the sum of shared and individual parameters. However, in \eqref{Eqn:MEMGeneral}, the function value itself is mixed. Both approaches are equivalent only if $g(\cdot)$ and $l^{(i)}(\cdot)$ are linear mappings, which is not the case in general. The graphical representation of the framework \eqref{Eqn:MEMGeneral} is shown in Figure \ref{PGM1}.

\subsection{Mixed Effect Model using GPs}\label{subsec:ME-GP}
As a concrete example of framework \eqref{Eqn:MEMGeneral}, we first consider the case where both $g(\cdot)$ and $l^{(i)}(\cdot)$ follow Gaussian Processes. Note that this formulation is just to relate our framework to existing multi-task GPs modeling each patient using a personalized GP. At the end of this subsection, it will be clear that this direction of individualization will involve almost intractable computations as the number of patients grows. Specifically, both components are represented as followings:
{\small
\begin{align*}
	g(\x_t) \sim \mathcal{GP} \big(0, k_g(\x_t, \x_{t'})\big), \enskip
	l^{(i)}(\x_t) \sim \mathcal{GP} \big(0, k^{(i)}(\x_t, \x_{t'})\big)
\end{align*}
}
where we assume both GPs to have zero-mean for simplicity, and $k_g(\cdot, \cdot)$ and $k^{(i)}(\cdot, \cdot)$ are valid covariance functions such as squared exponential kernel (RBF). We name this instantiation ME-GP that stands for Mixed Effect Model using GPs. Note that in this model, knowledge sharing occurs via the covariance function $k_g(\cdot, \cdot)$ of global GP. Further assuming the independence between $g(\cdot)$ and $l^{(i)}(\cdot)$ for all patients, we can derive overall covariance function in the following manner:
\begin{align*}
	\Tilde{k}(\x^{(i)}_t, \x^{(j)}_{t'}) = k_g(\x^{(i)}_t, \x^{(j)}_{t'}) + \delta_{ij} \cdot k^{(i)}(\x^{(i)}_t, \x^{(j)}_{t'})
\end{align*}
where $\delta_{ij}$ is the Kronecker delta function: $\delta_{ij}=1$ if $i=j$ (that is, for same patient) otherwise $0$. Interestingly, personalized GPs from this construction in fact boils down to a single GP with the covariance function $\Tilde{k}(\cdot, \cdot)$ for all of function variables $\f^{(1)}, \cdots, \f^{(P)}$:
{\small
\begin{align}\label{EQ:ME-GP}
\begin{bmatrix}
    \f^{(1)} \\
    \vdots \\
    \f^{(P)}
\end{bmatrix} 
& \sim \mathcal{GP}
\begin{pmatrix}
    \bm{0} ,
    \begin{bmatrix}
        K^g_{11} + K^{(1)} & \cdots & K^g_{1P} \\
        \vdots & \ddots & \vdots \\
        K^g_{P1} & \cdots & K^g_{PP} + K^{(P)}
    \end{bmatrix}
\end{pmatrix}
\end{align}
}
where $\f^{(i)}=f^{(i)}(\X_i)$ is a random vector of $i$-th patient process, and $K^g_{ij}$ and $K^{(i)}$ are covariance matrices with elements given by $k_g(\x^{(i)}_t, \x^{(j)}_{t'})$ and $k^{(i)}(\x^{(i)}_t, \x^{(i)}_{t'})$ respectively at $(t, t')$ position. As a result, the covariance matrix in~\eqref{EQ:ME-GP} lies in the space of $\mathcal{R}^{PT \times PT}$, assuming $T=T_i$ for all $i$. Given the model search and inference for a new point in GP rely on the inversion of covariance matrix, which costs $O(P^3T^3)$ for exact computation, learning with EHR datasets can be intractable in ME-GP even if we have small number of data points for each patient.

As noted earlier, Multi-task Gaussian Process (MTGP) is another model that uses GP in multi-task setting by forming covariance function with \emph{multiplicative} task-relatedness parameters as allows :
\begin{align*}
	\Tilde{k}(\x^{(i)}_t, \x^{(j)}_{t'}) = K_{ij} \cdot k_g(\x^{(i)}_t, \x^{(j)}_{t'})
\end{align*}
where $K_{ij}$ is an element at $(i, j)$ position in task-relatedness matrix $K$ as defined in \citep{mtgp2}. MTGP also requires to compute the inverse of huge covariance matrix in the same space as \eqref{EQ:ME-GP} and causes the same scalability issue.

\subsection{Deep Mixed Effect Model using GPs} \label{subsec:MECGP}
In this section, we propose Deep Mixed Effect Model using GPs (DME-GP) that exploits complementary properties of a deep network and GP and show our proposed model naturally overcomes scalability issue arisen in ME-GP. Specifically, we assume $g(\cdot)$ to be a deep network and $l^{(i)}$ to be GP as followings:
\begin{align*}
	g(\x_t) = \mu(\x_t), \enskip
	l^{(i)}(\x_t) \sim \mathcal{GP} \big(0, k^{(i)}(\x_t, \x_{t'})\big)
\end{align*}
where $\mu(\cdot)$ is any kind of a deep neural network such as MLP or RNN where the knowledge sharing occurs across individual processes of patients. As we have done in the previous subsection, we can derive overall covariance function as follow:
\begin{align*}
	\Tilde{k}(\x^{(i)}_t, \x^{(j)}_{t'}) = \delta_{ij} \cdot k^{(i)}(\x^{(i)}_t, \x^{(j)}_{t'}) \,\, .
\end{align*}
Note that this covariance function naturally forms block-diagonal matrix where each corresponds to each patient's process $\f^{(i)}$:
{\small
\begin{align}\label{EQ:DME-GP}
\begin{bmatrix}
    \f^{(1)} \\
    \vdots \\
    \f^{(P)}
\end{bmatrix} 
& \sim \mathcal{GP}
\begin{pmatrix}
    \begin{bmatrix}
        \bm{\mu}_1 \\
        \vdots \\
        \bm{\mu}_P
    \end{bmatrix}
     ,
    \begin{bmatrix}
        K^{(1)} & \cdots & \bm{0} \\
        \vdots & \ddots & \vdots \\
        \bm{0} & \cdots & K^{(P)}
    \end{bmatrix}
\end{pmatrix}
\end{align}
}
where $\bm{\mu}_i=\mu(\X_i)$ are outputs of a deep network. This in turn makes each patient process to be independent to other processes, which results in personalized GP models sharing global deep networks, described in~\eqref{Eqn:MEMGPRNN}. The computational cost of DME-GP compared to ME-GP reduces to $O(PT^3)$ thanks to its personalized formulation, which means DME-GP linearly scales to the number of patients $P$. 

We also investigate complementary properties of DME-GP between global and individual components. As we discussed in the introduction, shared function $g(\cdot)$ and individual function $l^{(i)}(\cdot)$ have their own desired properties:

\subsubsection{Individual Component} We adopt a personalized Gaussian Process for $l^{(i)}(\cdot)$.
This adoption allows the overall model to naturally provide the prediction uncertainty as a probabilistic model. In addition to that, GP enables us to reliably estimate individual signals based on relatively small number of data points (or visits for a patient) as a non-parametric model.

\subsubsection{Global Component} We adopt representationally expressive deep models such as MLP or RNN for $g(\cdot)$. This is a reasonable choice to capture complex patterns in high dimensional medical data in relatively computationally amenable fashion using stochastic gradient descent algorithms such as Adam \cite{adam}. 

\begin{figure}[t]
	\centering
	\includegraphics[width=1.0\linewidth]{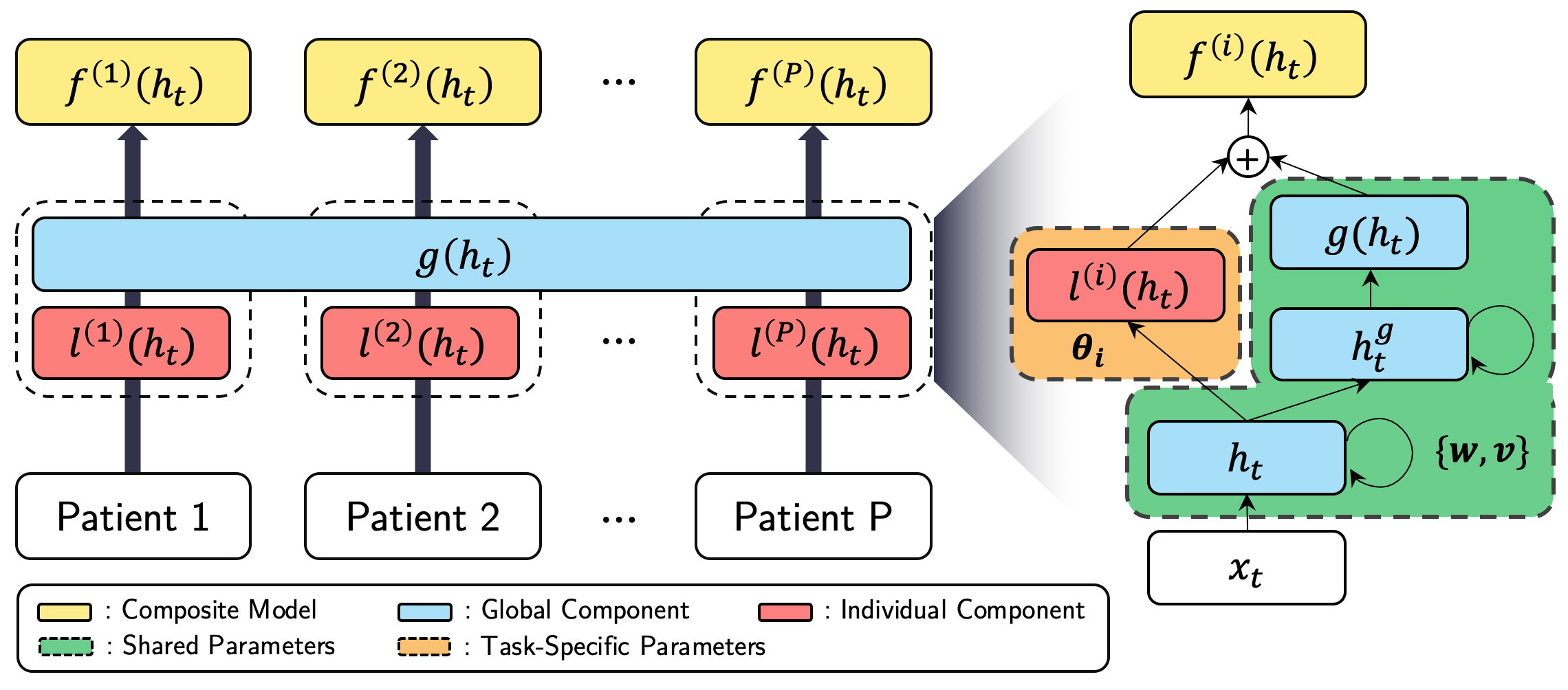}
	\vskip -.2cm
	\caption{\small An overall conceptual illustration of DME-GP. Left panel describes personalized formulation for each patient. Right panel shows detailed descriptions of decomposed components for a single patient. Note that weights sharing of deep model occurs across all GPs and individual parameters are maintained for them as shown in green and orange boxes respectively.}
	\label{FigConcept}
	\vspace{-.5cm}
\end{figure}

\subsubsection{Composite Model} Armed with these deliberate choices, it turns out the composite model \eqref{Eqn:MEMGeneral} can be reduced to personalized GPs \emph{sharing} a deep mean function, derived from~\eqref{EQ:DME-GP}:
\begin{align}\label{Eqn:MEMGPRNN}
	f^{(i)}(\x_t) \sim \mathcal{GP} \Big(\mu(\h_t|\w), k^{(i)}(\h_t, \h_{t'}|\t_i)\Big)
\end{align}
where the shared deep function $g(\cdot)$ is renamed as $\mu(\cdot|\w)$ (since it is a ``mean function'' of GP), $k^{(i)}(\cdot,\cdot|\t_i)$ is a kernel function for the individual process $l^{(i)}(\cdot)$, and $\h_t$ is some embedding for input $\x_t$ through global embedding function $\phi(\cdot|\v)$. Here we adopt deep models as global embedding function, as done for global component. Note, this is a natural extension to benefit from deep kernel approach to make local kernel function more expressive \citep{dkgp1}. The graphical representation of \eqref{Eqn:MEMGPRNN} is shown in Figure \ref{PGM2} although some parts of our model are deterministic mappings.

\subsubsection{Design Choice}
The framework of \eqref{Eqn:MEMGPRNN} does not restrict $\mu(\cdot)$ and $\phi(\cdot)$ to have specific form. However, we focus on RNN to efficiently handle sequential nature of EHR. For instance of vanilla RNN with single hidden layer case, we have:
\begin{align}\label{EqnH}
    \h_t = \phi(\x_t, \h_{t-1} | \v) = \tanh(\v_{xh}\x_t + \v_{hh}\h_{t-1})
\end{align}
where we suppress bias terms for simplicity and $\v=\{\v_{xh}, \v_{hh}\}$. $\mu(\cdot)$ can be formulated in a similar way. Note that the type of RNN cell can be any of choice, such as LSTM or GRU, and the architecture of a deep model can be carefully designed with domain knowledge of target dataset. Overall conceptual illustration of DME-GP is shown in Figure~\ref{FigConcept}.

\subsection{Learning and Inference of DME-GP}\label{sec:optim}
While our model is generally applicable to both regression and classification tasks, we implicitly assume the Gaussian likelihood throughout the paper just for clarity and notational simplicity. These can be seamlessly extended for classification problems with binary likelihood along with standard approximation techniques as in regular Gaussian Process.

\subsubsection{Learning}
Our learning objective is to maximize the marginal log-likelihoods of patients data $\mathcal{D}$ under the modeling assumption of \eqref{EQ:DME-GP} and \eqref{Eqn:MEMGPRNN} to find global-level parameters $\{ \w, \v \}$ and $\t = \{ \t_i \}_{i=1}^{P}$ from the individual components:
\begin{align}\label{Eqn:LL2}
	\t^*, \w^*, \v^* = \argmax_{\t, \w, \v} \sum_{i=1}^P \log p(\y_i | \X_i, \t, \w, \v)
\end{align}
where the log-likelihood is the sum of individual patient data under i.i.d. assumption across patients. 
An individual log-likelihood of single patient then can be represented using global and local parameters as follows:
{\small
\begin{align}\label{Eqn:LL3}
	\log p(\y_i | \X_i, \t_i, \w, \v) = & -\frac{1}{2} \big( \y_i - \bm{\mu}_i \big)^T {K^{(i)}}^{-1} \big( \y_i - \bm{\mu}_i \big) \nonumber \\
	&  - \frac{1}{2} \log |K^{(i)}| -\frac{T_i}{2} \log 2\pi
\end{align}
}
where the parameter dependencies are implicitly defined: $\bm{\mu}_i = [\mu(\h_1 | \w), \cdots, \mu(\h_{T_i} | \w)]^T$, $K^{(i)} \in \mathbb{R}^{T_i \times T_i}$ is a full covariance matrix given the element $k^{(i)}(\h_t, \h_{t'}|\t_i)$ at $(t, t')$ position, and RNN-based embedding $\h_t$ is a function on $\v$ as mentioned in \eqref{EqnH}.

The gradient of \eqref{Eqn:LL3} with respect to parameters can then be derived by chain rule as follows:
{\small
\begin{align}
	\frac{\partial \mathcal{L}_i}{\partial \t_i} &= \frac{\partial \mathcal{L}_i}{\partial K^{(i)}} \frac{\partial K^{(i)}}{\partial \t_i}  \,\, , \,\, \frac{\partial \mathcal{L}_i}{\partial \w} = \sum_{t=1}^{T_i} \frac{\partial \mathcal{L}_i}{\partial \mu_t} \frac{\partial \mu_t}{\partial \w} \nonumber \\
	\frac{\partial \mathcal{L}_i}{\partial \v} &= \frac{\partial \mathcal{L}_i}{\partial K^{(i)}} \sum_{t=1}^{T_i} \frac{\partial K^{(i)}}{\partial \h_t} \frac{\partial \h_t}{\partial \v} + \sum_{t=1}^{T_i} \frac{\partial \mathcal{L}_i}{\partial \mu_t} \sum_{t'=1}^{t} \frac{\partial \mu_t}{\partial \h_{t'}} \frac{\partial \h_{t'}}{\partial \v} \label{Eqn:DERI1_2}
\end{align}
}
where $\mathcal{L}_i := \log p(\y_i | \X_i, \t_i, \w, \v)$, and $\mu_t = \mu(\h_t)$. Note that, unlike vanilla RNN, the gradient computation of $\v$ from \eqref{Eqn:LL2} involves additional $\{K^{(i)}\}_{i=1}^{P}$ terms, leading to a bit more complicated computation. Note also that the gradient of global parameters $\w$ and $\v$ should involve the marginal likelihood across all patients while we only consider individual $\mathcal{L}_i$ for clarity. 

Our learning algorithm is based on stochastic gradient ascent in an alternating fashion and summarized in Appendix. Note again that our personalized formulation allows us to be able to avoid heavy computational cost from huge GP like in~\eqref{EQ:ME-GP} with EHR datasets. In addition, deep architectures as a shared mean function can be updated efficiently through the standard back-propagation algorithm. Note also that in non-Gaussian likelihood cases such as classification tasks, the marginal likelihood can be computed via variational lower bound with variational approximation or by simulation approaches \citep{approxGP}.

\subsubsection{Inference for new patient $j$}
Since we have single GP for each patient in \eqref{Eqn:MEMGPRNN}, our inference procedure for new patient $j$ follows the standard procedures of single GP inference. Suppose we want to predict $y_t$ of a new patient $j$ given current input feature $\x_t$ and all historical data on this patient: $\X=\{\x_s\}_{s=1}^{t-1}$ and $\y=\{y_s\}_{s=1}^{t-1}$ where we suppress the patient index $j$ for clarity. Then, we update the patient-specific parameters $\t_j$ of new GP by maximizing marginal log-likelihood \eqref{Eqn:LL3}, while global parameters $\{ \w, \v \}$ are fixed.  

 The predictive distribution of $y_t$ becomes $p(y_t | \x_t, \X, \y)=\mathcal{N}(y_t | \bar{y}_t, \sigma_t^2)$ with:
\begin{align}\label{Eqn:GPinference}
	\bar{y}_t &= \mu(\h_t) + \bm{k}_t^T K^{-1}(\y - \mu(\bm{H})) \nonumber \\
	\sigma_t^2 &= k(\h_t, \h_t) - \bm{k}_t^T K^{-1} \bm{k}_t
\end{align}
where $\bm{H}=[\h_1, ..., \h_{t-1}]^T$ and $\bm{k}_t = k(\bm{H}, \h_t)$.
The predictions can be done in sequential manner, which means we can predict the output at any time point of the patient. Note that the prediction at the first time point, $t=1$, can be done deterministically by the global mean function, where the model predicts in average. As we increase the time point $t$, we have more evidence for the patient and make better personalized predictions.

We note that approximate predictions can also be derived with non-Gaussian likelihood in a classification problem. While following the notations from Gaussian likelihood case explained above, the output $y$ follows some distribution $p\big(y|f(\x)\big)$ that is properly defined according to $\Y$ (i.e., normal distribution when $\Y:=\reals$ and Bernoulli distribution when $\Y:=\{0,1\}$). Then, the distribution of the latent function of GP for the test case $\x_t$ is given by:
\begin{align}\label{Eqn:GPf}
	p(f_t | \x_t, \X, \y) = \int p(f_t | \x_t, \X, \f) p(\f | \X, \y) d \f
\end{align}
where $p(\f | \X, \y) = p(\y | \f) p(\f | \X) / p(\y | \X)$ by \emph{Bayes' rule}.
Finally, the predictive distribution of $y_t$ is:
\begin{align}\label{Eqn:GPy}
	p(y_t | \x_t, \X, \y) = \int p(y_t|f_t) p(f_t | \x_t, \X, \y) df_t
\end{align}
where $p(y_t|f_t)$ is a properly designed likelihood function of $y_t$ given $f_t$ according to the class of problems. In regression case, we have analytic forms for \eqref{Eqn:GPf} and \eqref{Eqn:GPy} when $p(y_t|f_t)$ follows Gaussian as we have shown in~\eqref{Eqn:GPinference}. On the other hand for classification problems, the likelihood function is designed to be a sigmoid function such as $\frac{1}{1+\exp(-f_t)}$, which makes the integral in \eqref{Eqn:GPf} and \eqref{Eqn:GPy} analytically intractable. Thus, we need approximation methods for the posterior $p(\f|\X,\y)$, such as Laplace approximation, variational method, or Markov Chain Monte Carlo (MCMC) approximation \citep{approxGP,vgp}.

%%%%%%%%%%%%%%%%%%%%%%%%%%%%%%%%%%%%%%%%%%%%%%%%%%%%%%%%%%%%%%%%%%%%

\section{Experiments}\label{sec:experiments}

\subsection{Dataset Description and Experimental Setup}

\subsubsection{Vital-Sign Dataset}

\begin{figure*}[t]
    \centering
    \vspace{-.2cm}
    \includegraphics[width=0.85\linewidth]{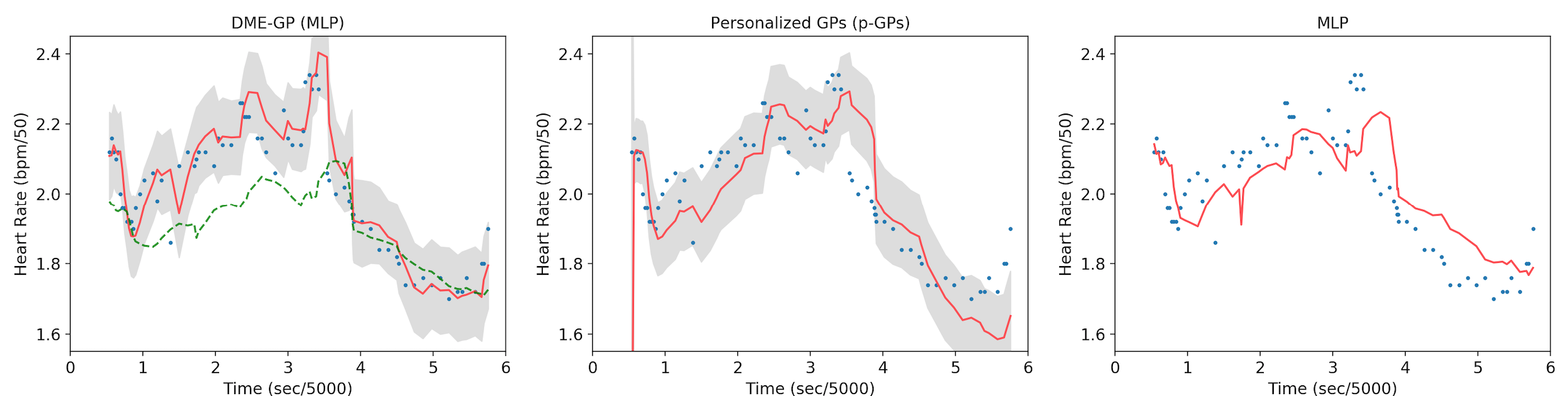}
    \vskip -.1cm
    \caption{\small \textbf{Vital-Sign Analysis.} The predictions (red curves) for a random patient (blue dots) are shown in the order of DME-GP, p-GPs, and MLP respectively. The uncertainty representation is given by $\pm$ 1 standard deviation centered at the model's predictions. The global trend of DME-GP, which is predicted by a global mean function, is shown as a green dashed line.}
    \label{VSA}
    \vspace{-.5cm}
\end{figure*} 

This dataset is compiled from a publicly available EHR dataset called Physionet Challenge 2012 \citep{physionet}. Specifically, we extract heart rate (HR) information in time-series for 865 patients who are in a cardiac surgery recovery unit (52,942 overall data points). Input (event time) and output values (heart rate) are scaled by 5,000 and 50, respectively. The dataset is motivated from the patient monitoring system in hospitals where the conventional procedures of the system are operated by nursing staff who frequently check target vital signs. It is important to automate the monitoring system to reduce the high cost of human labor and for early detection of those patients in a dangerous condition by predicting progressions of vital signs.

\subsubsection{Medical Checkup Dataset}
This dataset is compiled from health checkup records for 32,927 patients (and 220,408 data points) collected from 2002 to 2013 (provided by National Health Insurance Service; NHIS). We select 12 common target diseases, and for each disease we have the health checkup history (either real or categorical input features) and corresponding target binary variables indicating either the absence or presence of a disease at each year. We convert categorical variables into one-hot vectors and normalize each real-valued feature with its mean and standard deviation. We simply fill in missing values in raw EHR data with zeros since missing rate is low \citep{rnn3,rnn2}. Detailed statistics of our data for each dataset are provided in Appendix.

\subsubsection{Baselines}
We compare DME-GP against several baseline models including RNN-based deep models that are known to work well with time-series datasets:

\vspace{-0.1cm}
\begin{itemize}
\item \textbf{Linear Models (LM):} A linear and logistic regression models for regression and classification respectively.
\vspace{-0.1cm}
\item \textbf{MLP:} A multi-layer perceptron containing two hidden layers with a sigmoid activation function.
\vspace{-0.1cm}
\item \textbf{RNN:} A RNN containing two hidden layers with long short-term memory units (LSTM) \citep{rnn1}.
\vspace{-0.1cm}
\item \textbf{RETAIN:} A RNN-based recurrent attention model proposed in~\citet{retain}.
\vspace{-0.1cm}
\item \textbf{MTGP-RNN:} A multi-task Gaussian Process-wrapped RNN proposed in~\citet{sepsis}. This model uses a multi-task GP (MTGP), but it is computationally tractable since it only considers the MTGP to correlate input features across different time points. 
\vspace{-0.1cm}
\item \textbf{MAML:} A RNN-based Meta-SGD model proposed in~\citet{meta-sgd}. This model is extended version of MAML~\citep{maml} where the model also learns step size of a meta-learner (an optimizer).
\end{itemize}
\vspace{-0.1cm}
In case of LM and MLP, we treat individual time steps for all patients as i.i.d. observations since they are not specifically designed for time-series inputs. 
We exclude comparisons against variants of a single GP including MTGP and ME-GP not only because we observed their limited performances on the preliminary experiments but because they are computationally too expensive for exact inference. 
For our DME-GP, we consider two different models that use MLP and RNN with one hidden layer respectively. 
Note that we use a single-layered deep kernel function for our DME-GPs for fair comparisons (since baseline deep models use two layers in total). We defer all training details (e.g. setting hyper-parameters) to Appendix due to the space constraint. 

\begin{table*}
\vspace{-.2cm}
    \begin{minipage}{0.7\linewidth}
        \caption{\small \textbf{Disease Risk Prediction.} Performance (AUC) comparisons for 12 risk prediction tasks.}
        \label{BASELINES}
        \vspace{-.2cm}
        \centering
\scalebox{0.72}{
\begin{tabular}{lcccccccc}
\toprule
\multirow{2}{*}{Diseases} & DME-GP & DME-GP & \multirow{2}{*}{RNN} & \multirow{2}{*}{RETAIN} & \multirow{2}{*}{MTGP-RNN} & \multirow{2}{*}{MLP} & \multirow{2}{*}{MAML} & \multirow{2}{*}{LM} \\
& (RNN) & (MLP) &&&&&& \\
\midrule
Alcoholic Fatty Liver &$\bm{0.829}$&0.801&0.791&0.796&0.785&0.777&0.780&0.529\\
Atherosclerosis           &$\bm{0.815}$&0.740&0.662&0.726&0.716&0.735&0.728&0.547\\
Emphysema             &$\bm{0.805}$&0.742&0.778&0.671&0.769&0.787&0.632&0.552\\
Liver Cirrhosis           &$\bm{0.932}$&0.922&0.888&0.871&0.904&0.856&0.913&0.635\\
Alcoholic Hepatitis       &0.842&0.852&0.803&0.788&$\bm{0.853}$&0.782&0.852&0.563\\
Arrhythmia                &0.763&0.740&0.592&0.616&0.587&0.658&$\bm{0.767}$&0.602\\
Fatty Liver               &0.726&$\bm{0.731}$&0.689&0.684&0.680&0.647&0.691&0.513\\
Heart Failure             &$\bm{0.829}$&0.759&0.790&0.792&0.761&0.783&0.729&0.620\\
Hepatic Failure       &0.728&$\bm{0.738}$&0.625&0.614&0.646&0.688&0.653&0.563\\
Hepatitis B           &0.542&0.489&0.567&0.571&$\bm{0.674}$&0.671&0.554&0.528\\
Myocardial Infarction     &0.885&0.826&0.865&0.858&0.815&$\bm{0.890}$&0.803&0.787\\
Toxic Liver Disease   &0.641&$\bm{0.698}$&0.595&0.594&0.596&0.643&0.685&0.518\\
\midrule
Task Average &$\bm{0.778}$&0.753&0.720&0.715&0.732&0.743&0.732&0.580 \\
\bottomrule
\end{tabular}}
    \end{minipage}
    \hfill
    \begin{minipage}{0.3\linewidth}
        \centering
        \includegraphics[width=1.0\linewidth]{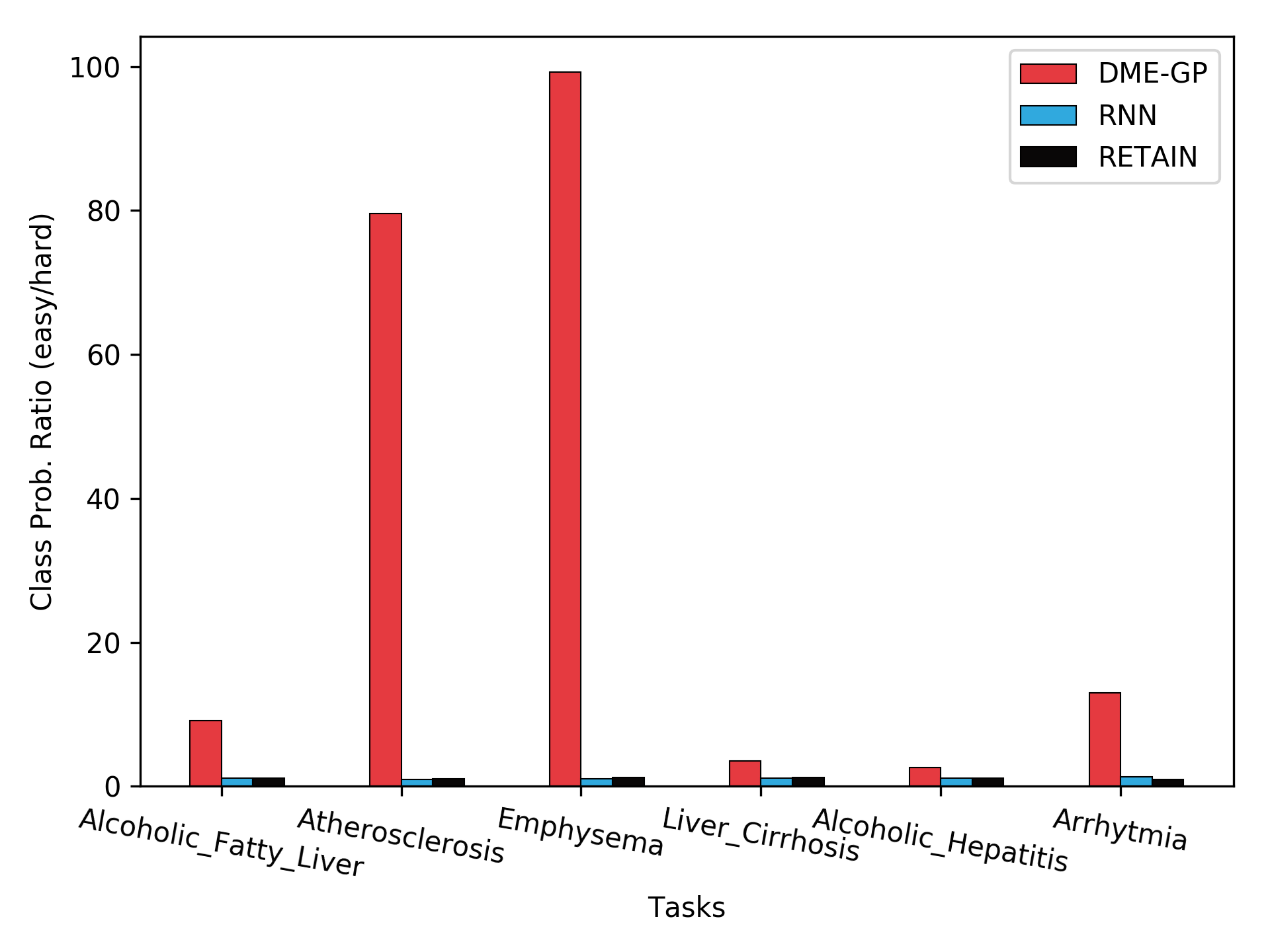}
        \captionof{figure}{\small \textbf{Reliability Study.} Average classification probability ratio between easy patients and hard patients.}
        \label{FigCali}
    \end{minipage}
    \vspace{-.4cm}
\end{table*}

\subsubsection{Ablation Models}
We also evaluate the following variants of DME-GP for an ablation study:

\vspace{-0.1cm}
\begin{itemize}
\item \textbf{p-GPs:} Personalized GPs with zero mean, individual embedding $\v_i$ and covariance $\t_i$ for patient-$i$.
\vspace{-0.1cm}
\item \textbf{p-GPs-cov:} Personalized GPs with zero mean, shared embedding $\v$ and covariance $\t$. 
\vspace{-0.1cm}
\item \textbf{p-GPs-both:} Personalized GPs with shared mean and covariance parameters $\w, \v, \t$.
\end{itemize}
\vspace{-0.1cm}
We expect that p-GPs would not generalize well on a relatively small amount of patient data due to lack of sharing information. p-GPs-cov would benefit sharing information from a shared covariance function but in a limited way and lose individual characteristics. Lastly, p-GPs-both would fully benefit from both shared mean and covariance function, but not be able to capture individual signals because of missing local components. Interestingly, \citep{pgpboth} proposes a similar model as p-GPs-both concurrently to our work but in meta-learning perspective.

The code is available at \textit{https://github.com/jik0730/Deep-Mixed-Effect-Model-using-Gaussian-Processes}.

\subsection{Vital-Sign Analysis: Heart Rate}\label{exp:VSA}

Our goal here is to find a mapping from a fixed window time-series $\{ x_{t-2:t}, y_{t-2:t} \}$ to step-ahead target value $y_{t+5}$ at every time stamp $t$. 

In this experiment, we compare DME-GP (MLP) against p-GPs and MLP since other baselines perform similarly with these two baselines. Running examples made by these models for a selected patient are shown in Figure~\ref{VSA}. p-GPs shown in the middle graph tends to produce underestimated predictions where the model outputs lower values than expected, especially in initial time points. This phenomenon can be explained by its lack of global trend, and we can verify the benefit of knowledge transferring from other patients.

MLP on the right tends to behave like a \emph{follower} where the predictions simply copy the former time-series targets since previous targets are the most useful information for population based models. On the other hand, DME-GP shows better predictions than the baselines. The predicted global trend (or mean) in DME-GP (shown in green dashed line) exhibits a similar pattern with the predictions of MLP and contributes to making the overall predictions better than p-GPs and MLP. This result partially implies that DME-GP is able to successfully benefit from both global and individual components. Overall test prediction performance (RMSE) for all patients is measured as 0.150 (DME-GP), 0.243 (p-GPs), and 0.194 (MLP), respectively.

\begin{figure}[t]
	\centering
	\vspace{-.2cm}
	\includegraphics[width=1.0\linewidth]{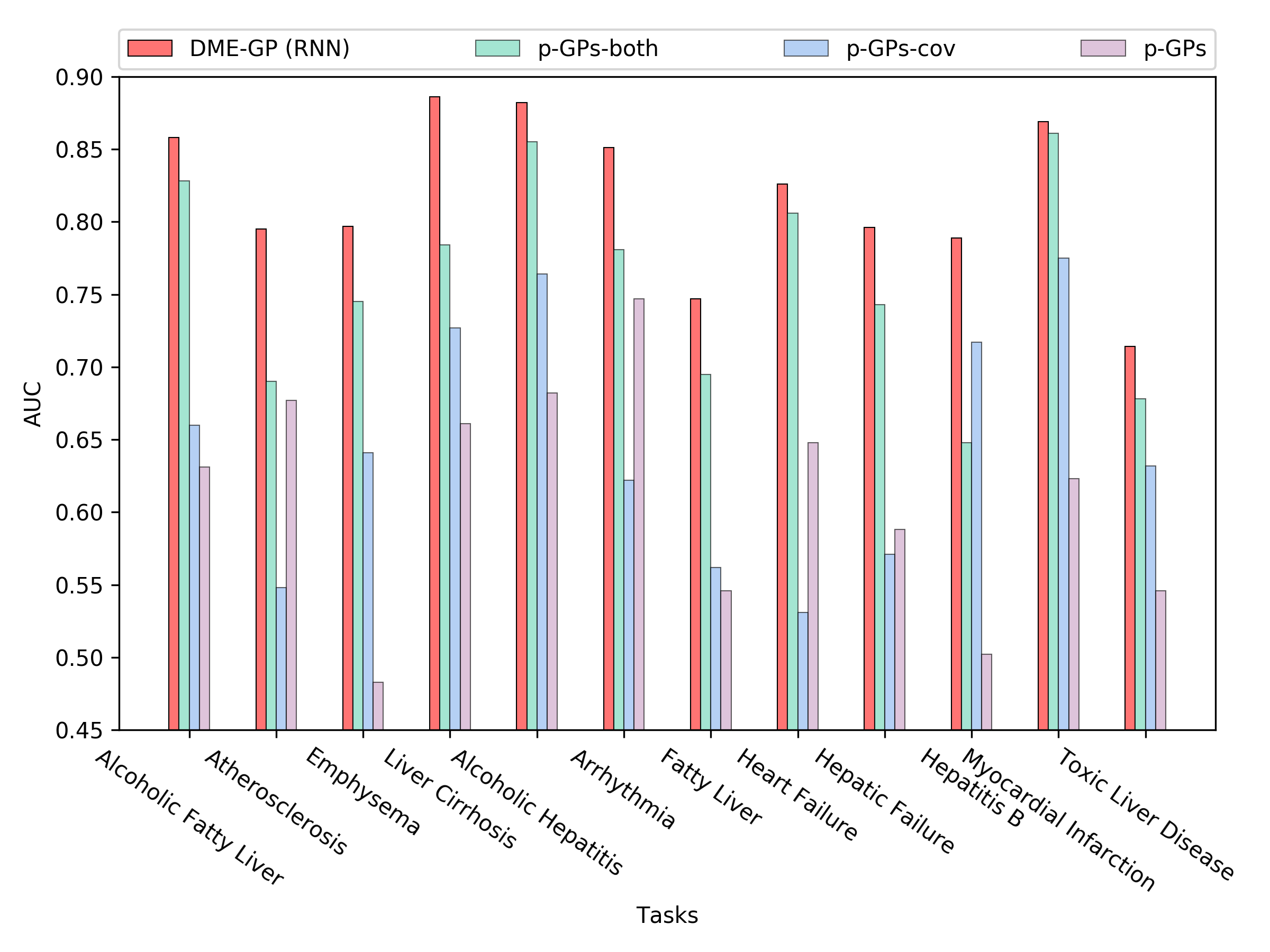}
	\vskip -.3cm
	\caption{\small \textbf{Ablation Study.} Performance comparisons (in terms of Area Under the ROC Curve; AUC) among the variants of DME-GP. For fair comparison, we evaluate the models under the same hyper-parameters and measure the validation AUC.}
	\label{VARIANTS}
	\vspace{-.5cm}
\end{figure}

\subsection{Disease Risk Prediction by Medical Checkup}\label{exp:DRP}

Given a visit sequence of input features $\{ \x^{(i)}_t \}_{t=1}^{T_i}$ (which are representing clinical status) and corresponding binary targets $\{y^{(i)}_t \}_{t=1}^{T_i-1}$ (disease history) for each patient $i$, our task is to predict the most recent target $y^{(i)}_{T_i}$. The task can be thought of as predicting the risk of disease given time-series health checkup variables.

We compare DME-GP against standard baselines listed above to verify the importance of considering the idiosyncratic variability of individual patient when modeling heterogeneous clinical data. As summarized in Table~\ref{BASELINES}, DME-GP significantly outperforms others in most of the cases. The performance of DME-GP with a RNN mean function is the best among them probably because RNN is able to effectively capture the global trend by making use of historical data points in time-series. The population-based deep models such as RNN perform worse than DME-GP probably because they are not be able to effectively model individual differences across diverse patients. In particular, MTGP-RNN's degraded performance compared to DME-GP suggests that modeling each \emph{patient} as a single task is better than modeling each \emph{feature} when modeling heterogeneous patient data. MAML also shows not enough prediction scores compared to DME-GP, which supports the claim that DME-GP is a better way to transfer knowledge in heterogeneous EHR datasets.

\subsubsection{Ablation Study}
In this experiment, we evaluate the variants of DME-GP to investigate the effect of using different levels of information sharing across patients. As a recap, our model only shares the global \emph{mean function} across all patient-wise GPs to capture global trend in the data for knowledge transfer, and leverages patient-wise GPs to capture local variability from inherent hidden factors of each patient. The results shown in Figure~\ref{VARIANTS} support our claim that the decomposition into a shared global part and a personalized local part is sensible with heterogeneous medical data. 

\begin{table}[t]
\caption{\small \textbf{Reliability Study.} Performance (in terms of Area Under the ROC Curve; AUC) comparisons when we exclude a set of \emph{hard patient} who has a positive label at $T_i$ but negative labels for others. Experiments for other diseases are shown in Appendix.}
\vspace{-.3cm}
\label{cali}
\begin{center}
\resizebox{7.0cm}{!}{
\begin{sc}
\begin{tabular}{lccc}
\toprule
Diseases & DME-GP & RNN & RETAIN \\
\midrule
A. Fatty Liver		& $\bm{0.998}$	& 0.856	& 0.861 \\
Atherosclerosis		& $\bm{1}$	& 0.683	& 0.768 \\
Emphysema    		& $\bm{0.991}$	& 0.785	& 0.765 \\
Liver Cirrhosis   	& $\bm{0.989}$	& 0.928	& 0.922 \\
Alcoholic Hepatitis   	& $\bm{0.981}$	& 0.881	& 0.866 \\
Arrhythmia     		& $\bm{0.992}$	& 0.639	& 0.656 \\
\bottomrule
\end{tabular}
\end{sc}
}
\end{center}
\vspace{-.5cm}
\end{table}

\subsubsection{Reliability Study}
Finally, in order to indirectly measure the predictive reliability of our model, we design a simple modification from the previous experiment on risk predictions. Specifically, we define a \emph{hard patient} to denote a patient who has a positive label \emph{only} at the prediction time $T_i$ but never has positive labels in his/her historical data. We compare the differences (in terms of confidence as well as AUC) between i) the case where we exclude hard patients (easy) and ii) the case where we only consider such hard patients (hard). Figure \ref{FigCali} shows how confident the models are for two groups of patients and Table \ref{cali} summarizes AUC when we exclude hard patients. Our model exhibits clear distinctions between the two cases and achieves almost perfect scores for many datasets when it is confident. On the other hand, for the latter case (only on hard patients), the gain of using our model degrades as shown in Appendix, while being competitive with deep models. This reliable confidence information will allow proper involvement of human medical staff.

%%%%%%%%%%%%%%%%%%%%%%%%%%%%%%%%%%%%%%%%%%%%%%%%%%%%%%%%%%%%%%%%%%%%

\section{Conclusion}\label{sec:conclusion}

We have presented the framework of Mixed Effect Model for electronic health records (EHR) and provided Deep Mixed Effect Model using GPs (DME-GP) as a showcase example that exploits complementary properties of RNN and GP. In DME-GP, we use a deep network to learn a globally shared mean function capturing complex global patterns among diverse patients and use GP to build personalized and reliable prediction model. We have investigated the properties of our model for diverse tasks complied from real EHR data and validated the superiority of it against state-of-the-art baselines. One last important note is that our model has an advantage to provide prediction uncertainty via GP in a principled way, which is essential for safety-critical clinical tasks.

%%%%%%%%%%%%%%%%%%%%%%%%%%%%%%%%%%%%%%%%%%%%%%%%%%%%%%%%%%%%%%%%%%%%

{\small
\section{Acknowledgments}
 This work was supported by the National Research Foundation of Korea (NRF) grants (No.2018R1A5A1059921, No.2019R1C1C1009192), Institute of Information \& Communications Technology Planning \& Evaluation (IITP) grants (No.2016-0-00563, No.2017-0-01779, No.2019-0-01371) funded by the Korea government (MSIT) and Samsung Research Funding \& Incubation Center via SRFC-IT1702-15.
}

%%%%%%%%%%%%%%%%%%%%%%%%%%%%%%%%%%%%%%%%%%%%%%%%%%%%%%%%%%%%%%%%%%%%

\bibliographystyle{aaai}
{\small
\bibliography{aaai2020}}

\begin{thebibliography}{}

\bibitem[\protect\citeauthoryear{Alaa \bgroup et al\mbox.\egroup
  }{2016}]{hete2}
Alaa, A.~M.; Yoon, J.; Hu, S.; and {van der Schaar}, M.
\newblock 2016.
\newblock Personalized risk scoring for critical care patients using mixtures
  of gaussian process experts.
\newblock In {\em ICML Workshop on Computational Frameworks for
  Personalization}.

\bibitem[\protect\citeauthoryear{Bonilla, Agakov, and Williams}{2007}]{mtgp4}
Bonilla, E.~V.; Agakov, F.~V.; and Williams, C. K.~I.
\newblock 2007.
\newblock Kernel multi-task learning using task-specific features.
\newblock In {\em AISTATS}.

\bibitem[\protect\citeauthoryear{Bonilla, Chai, and Williams}{2008}]{mtgp2}
Bonilla, E.~V.; Chai, K. M.~A.; and Williams, C. K.~I.
\newblock 2008.
\newblock Multi-task {G}aussian process prediction.
\newblock In {\em NIPS}.

\bibitem[\protect\citeauthoryear{Che \bgroup et al\mbox.\egroup }{2018}]{rnn2}
Che, Z.; Purushotham, S.; Cho, K.; Sontag, D.; and Liu, Y.
\newblock 2018.
\newblock Recurrent neural networks for multivariate time series with missing
  values.
\newblock {\em Scientific Reports} 8(1):6085.

\bibitem[\protect\citeauthoryear{Cheng \bgroup et al\mbox.\egroup
  }{2017}]{mtgp3}
Cheng, L.; Darnell, G.; Chivers, C.; Draugelis, M.~E.; Li, K.; and Engelhardt,
  B.~E.
\newblock 2017.
\newblock Sparse multi-output {G}aussian processes for medical time series
  prediction.
\newblock {\em arXiv preprint arXiv:1703.09112}.

\bibitem[\protect\citeauthoryear{Cho \bgroup et al\mbox.\egroup }{2014}]{gru}
Cho, K.; Merrienboer, B.; Gulcehre, C.; Bahdanau, D.; Bougares, F.; Schwenk,
  H.; and Bengio, Y.
\newblock 2014.
\newblock Learning phrase representations using {RNN} encoder-decoder for
  statistical machine translation.
\newblock In {\em EMNLP}.

\bibitem[\protect\citeauthoryear{Choi \bgroup et al\mbox.\egroup
  }{2016a}]{doctorai}
Choi, E.; Bahadori, M.~T.; Stewart, W.; and Sun, J.
\newblock 2016a.
\newblock {Doctor AI}: Predicting clinical events via recurrent neural
  networks.
\newblock In {\em Machine Learning for Healthcare}.

\bibitem[\protect\citeauthoryear{Choi \bgroup et al\mbox.\egroup
  }{2016b}]{retain}
Choi, E.; Bahadori, M.~T.; Sun, J.; Kulas, J.; Schuetz, A.; and Stewart, W.
\newblock 2016b.
\newblock {RETAIN}: An interpretable predictive model for healthcare using
  reverse time attention mechanism.
\newblock In {\em NIPS}.

\bibitem[\protect\citeauthoryear{Clifton \bgroup et al\mbox.\egroup
  }{2013}]{pgp1}
Clifton, L.; Clifton, D.~A.; Pimentel, M. A.~F.; Watkinson, P.~J.; and
  Tarassenko, L.
\newblock 2013.
\newblock {G}aussian processes for personalized e-health monitoring with
  wearable sensors.
\newblock {\em IEEE Transactions on Biomedical Engineering} 60(1).

\bibitem[\protect\citeauthoryear{Damianou and Lawrence}{2013}]{dgp1}
Damianou, A.~C., and Lawrence, N.~D.
\newblock 2013.
\newblock Deep {G}aussian processes.
\newblock In {\em AISTATS}.

\bibitem[\protect\citeauthoryear{D{\"u}richen \bgroup et al\mbox.\egroup
  }{2015}]{mtgp1}
D{\"u}richen, R.; Pimentel, M. A.~F.; Clifton, L.; Schweikard, A.; and Clifton,
  D.~A.
\newblock 2015.
\newblock Multitask {G}aussian processes for multivariate physiological
  time-series analysis.
\newblock {\em IEEE Transactions on Biomedical Engineering} 61(1).

\bibitem[\protect\citeauthoryear{Finn, Abbeel, and Levine}{2017}]{maml}
Finn, C.; Abbeel, P.; and Levine, S.
\newblock 2017.
\newblock Model-{A}gnostic {M}eta-{L}earning for fast adaptation of deep
  networks.
\newblock In {\em ICML}.

\bibitem[\protect\citeauthoryear{Fortuin and Ratsch}{2019}]{pgpboth}
Fortuin, V., and Ratsch, G.
\newblock 2019.
\newblock Deep mean functions for meta-learning in {G}aussian processes.
\newblock {\em arXiv preprint arXiv:1901.08098}.

\bibitem[\protect\citeauthoryear{Futoma \bgroup et al\mbox.\egroup
  }{2016}]{shared-gp2}
Futoma, J.; Sendak, M.; Cameron, B.; and Heller, K.
\newblock 2016.
\newblock Predicting disease progression with a model for multivariate
  longitudinal clinical data.
\newblock In {\em Machine Learning for Healthcare}.

\bibitem[\protect\citeauthoryear{Futoma \bgroup et al\mbox.\egroup
  }{2017}]{sepsis2}
Futoma, J.; Hariharan, S.; Sendak, M.; Brajer, N.; Clement, M.; Bedoya, A.;
  {O'}Brien, C.; and Heller, K.
\newblock 2017.
\newblock An improved multi-output {G}aussian process {RNN} with real-time
  validation for early sepsis detection.
\newblock In {\em Machine Learning for Healthcare}.

\bibitem[\protect\citeauthoryear{Futoma, Hariharan, and Heller}{2017}]{sepsis}
Futoma, J.; Hariharan, S.; and Heller, K.
\newblock 2017.
\newblock Learning to detect sepsis with a multitask {G}aussian process {RNN}
  classifier.
\newblock In {\em ICML}.

\bibitem[\protect\citeauthoryear{Gal and Ghahramani}{2016}]{Gal2016Dropout}
Gal, Y., and Ghahramani, Z.
\newblock 2016.
\newblock Dropout as a {B}ayesian approximation: Representing model uncertainty
  in deep learning.
\newblock In {\em ICML}.

\bibitem[\protect\citeauthoryear{Goldberger \bgroup et al\mbox.\egroup
  }{2000}]{physionet}
Goldberger, A.~L.; Amaral, A.~N.; Glass, L.; Hausdorff, J.~M.; Ivanov, P.;
  Mark, R.~G.; Mietus, J.~E.; Moody, G.~B.; Peng, C.; and Stanley, H.~E.
\newblock 2000.
\newblock {P}hysiobank, {P}hysiotoolkit, and {P}hysionet: {C}omponents of a new
  research resource for complex physiologic signals.
\newblock {\em Circulation} 100(23):e215--e220.

\bibitem[\protect\citeauthoryear{Greene}{2003}]{mixed-effect}
Greene, W.
\newblock 2003.
\newblock {\em Econometric Analysis}.
\newblock Pearson Education.

\bibitem[\protect\citeauthoryear{Hensman, Matthews, and
  Ghahramani}{2015}]{svgp}
Hensman, J.; Matthews, A. G. d.~G.; and Ghahramani, Z.
\newblock 2015.
\newblock Scalable variational {G}aussian process classification.
\newblock In {\em AISTATS}.

\bibitem[\protect\citeauthoryear{Hochreiter and Schmidhuber}{1997}]{lstm}
Hochreiter, S., and Schmidhuber, J.
\newblock 1997.
\newblock Long short-term memory.
\newblock {\em Neural Computation} 9(8):1735--1780.

\bibitem[\protect\citeauthoryear{Huang \bgroup et al\mbox.\egroup
  }{2015}]{dkgp3}
Huang, W.; Zhao, D.; Sun, F.; Liu, H.; and Chang, E.
\newblock 2015.
\newblock Scalable {G}aussian process regression using deep neural networks.
\newblock In {\em IJCAI}.

\bibitem[\protect\citeauthoryear{Iwata and Ghahramani}{2017}]{mean-nn}
Iwata, T., and Ghahramani, Z.
\newblock 2017.
\newblock Improving output uncertainty estimation and generalization in deep
  learning via neural network {G}aussian processes.
\newblock In {\em arXiv preprint arXiv:1707.05922}.

\bibitem[\protect\citeauthoryear{Jacobs \bgroup et al\mbox.\egroup
  }{1991}]{moe1}
Jacobs, R.~A.; Jordan, M.~I.; Nowlan, S.~J.; and Hinton, G.~E.
\newblock 1991.
\newblock Adaptive mixtures of local experts.
\newblock {\em Neural Computing}.

\bibitem[\protect\citeauthoryear{Kingma and Ba}{2015}]{adam}
Kingma, D.~P., and Ba, J.
\newblock 2015.
\newblock Adam: A method for stochastic optimization.
\newblock In {\em ICLR}.

\bibitem[\protect\citeauthoryear{Li \bgroup et al\mbox.\egroup
  }{2017}]{meta-sgd}
Li, Z.; Zhou, F.; Chen, F.; and Li, H.
\newblock 2017.
\newblock Meta-{SGD}: Learning to learn quickly for few-shot learning.
\newblock In {\em arXiv preprint arXiv:1707.09835}.

\bibitem[\protect\citeauthoryear{Lipton \bgroup et al\mbox.\egroup
  }{2016}]{rnn1}
Lipton, Z.~C.; Kale, D.~C.; Elkan, C.; and Wetzel, R.
\newblock 2016.
\newblock Learning to diagnose with {LSTM} recurrent neural networks.
\newblock In {\em ICLR}.

\bibitem[\protect\citeauthoryear{Lipton, Kale, and Wetzel}{2016}]{rnn3}
Lipton, Z.~C.; Kale, D.~C.; and Wetzel, R.
\newblock 2016.
\newblock Modeling missing data in clinical time series with {RNN}s.
\newblock In {\em Machine Learning for Healthcare}.

\bibitem[\protect\citeauthoryear{Marlin \bgroup et al\mbox.\egroup
  }{2012}]{hete3}
Marlin, B.~M.; Kale, D.~C.; Khemani, R.~G.; and Wetzel, R.~C.
\newblock 2012.
\newblock Unsupervised pattern discovery in electronic health care data using
  probabilistic clustering models.
\newblock In {\em Proceedings of the 2Nd ACM SIGHIT International Health
  Informatics Symposium}.

\bibitem[\protect\citeauthoryear{Micchelli, Xu, and Zhang}{2006}]{kernel1}
Micchelli, C.~A.; Xu, Y.; and Zhang, H.
\newblock 2006.
\newblock Universal kernels.
\newblock {\em JMLR} 7.

\bibitem[\protect\citeauthoryear{Ng \bgroup et al\mbox.\egroup }{2015}]{hete1}
Ng, K.; Sun, J.; Hu, J.; and Wang, F.
\newblock 2015.
\newblock Personalized predictive modeling and risk factor identification using
  patient similarity.
\newblock In {\em AMIA}.

\bibitem[\protect\citeauthoryear{Nguyen and Bonilla}{2014}]{mtgp5}
Nguyen, T.~V., and Bonilla, E.~V.
\newblock 2014.
\newblock Collaborative multi-output {G}aussian {P}rocesses.
\newblock In {\em UAI}.

\bibitem[\protect\citeauthoryear{Nickisch and Rasmussen}{2008}]{approxGP}
Nickisch, H., and Rasmussen, C.~E.
\newblock 2008.
\newblock Approximations for binary {G}aussian process classification.
\newblock {\em JMLR} 9.

\bibitem[\protect\citeauthoryear{Opper and Archambeau}{2009}]{vgp}
Opper, M., and Archambeau, C.
\newblock 2009.
\newblock The variational {G}aussian approximation revisited.
\newblock {\em Neural Computation} 21(3):786--792.

\bibitem[\protect\citeauthoryear{Peterson \bgroup et al\mbox.\egroup
  }{2017}]{pgp2}
Peterson, K.; Rudovic, O.; Guerrero, R.; and Picard, R.
\newblock 2017.
\newblock Personalized {G}aussian processes for future prediction of
  alzheimer’s disease progression.
\newblock In {\em NIPS Workshop on Machine Learning for Healthcare}.

\bibitem[\protect\citeauthoryear{Ravi and Larochelle}{2017}]{meta-lstm}
Ravi, S., and Larochelle, H.
\newblock 2017.
\newblock Optimization as a model for few-shot learning.
\newblock In {\em ICLR}.

\bibitem[\protect\citeauthoryear{Saito and Rehnmsmeier}{2015}]{roc1}
Saito, T., and Rehnmsmeier, M.
\newblock 2015.
\newblock The precision-recall plot is more informative than the {ROC} plot
  when evaluating binary classifiers on imbalanced datasets.
\newblock In {\em PLOS}.

\bibitem[\protect\citeauthoryear{Salakhutdinov and Hinton}{2007}]{dkgp2}
Salakhutdinov, R., and Hinton, G.
\newblock 2007.
\newblock Using deep belief nets to learn covariance kernels for {G}aussian
  processes.
\newblock In {\em NIPS}.

\bibitem[\protect\citeauthoryear{Salimbeni and Deisenroth}{2017}]{dgp2}
Salimbeni, H., and Deisenroth, M.~P.
\newblock 2017.
\newblock Doubly stochastic variational inference for deep {G}aussian
  processes.
\newblock In {\em NIPS}.

\bibitem[\protect\citeauthoryear{Schulam and Saria}{2015}]{shared-gp1}
Schulam, P., and Saria, S.
\newblock 2015.
\newblock A framework for individualizing predictions of disease trajectories
  by exploiting multi-resolution structure.
\newblock In {\em NIPS}.

\bibitem[\protect\citeauthoryear{Snell, Swersky, and Zemel}{2017}]{protonet}
Snell, J.; Swersky, K.; and Zemel, R.~S.
\newblock 2017.
\newblock Prototypical networks for few-shot learning.
\newblock In {\em NIPS}.

\bibitem[\protect\citeauthoryear{Snelson and Ghahramani}{2006}]{spgp}
Snelson, E., and Ghahramani, Z.
\newblock 2006.
\newblock Sparse {G}aussian processes using pseudo-inputs.
\newblock In {\em NIPS}.

\bibitem[\protect\citeauthoryear{Srivastava \bgroup et al\mbox.\egroup
  }{2014}]{dropout}
Srivastava, N.; Hinton, G.; Krizhevsky, A.; Sutskever, I.; and Salakhutdinov,
  R.
\newblock 2014.
\newblock Dropout: A simple way to prevent neural networks from overfitting.
\newblock {\em JMLR} 15(1).

\bibitem[\protect\citeauthoryear{Sung \bgroup et al\mbox.\egroup
  }{2018}]{relnet}
Sung, F.; Yang, Y.; Zhang, L.; Xiang, T.; Torr, P. H.~S.; and Hospedales, T.~M.
\newblock 2018.
\newblock Learning to {C}ompare: Relation network for few-shot learning.
\newblock In {\em CVPR}.

\bibitem[\protect\citeauthoryear{Swamidass \bgroup et al\mbox.\egroup
  }{2010}]{roc2}
Swamidass, S.~J.; A., A.~C.; K., D.; and Baldi, P.
\newblock 2010.
\newblock A {CROC} stronger than {ROC}: measuring, visualizing and optimizing
  early retrieval.
\newblock In {\em ISCB}.

\bibitem[\protect\citeauthoryear{Titsias}{2009}]{spgp2}
Titsias, M.~K.
\newblock 2009.
\newblock Variational learning of inducing variables in sparse {G}aussian
  processes.
\newblock In {\em AISTATS}.

\bibitem[\protect\citeauthoryear{Vinyals \bgroup et al\mbox.\egroup
  }{2016}]{matchingnet}
Vinyals, O.; Blundell, C.; Lillicrap, T.; Kavukcuoglu, K.; and Wierstra, D.
\newblock 2016.
\newblock Matching networks for one shot learning.
\newblock In {\em NIPS}.

\bibitem[\protect\citeauthoryear{Wilson \bgroup et al\mbox.\egroup
  }{2016}]{dkgp1}
Wilson, A.~G.; Hu, Z.; Salakhutdinov, R.; and Xing, E.~P.
\newblock 2016.
\newblock Deep kernel learning.
\newblock In {\em AISTATS}.

\end{thebibliography}

%%%%%%%%%%%%%%%%%%%%%%%%%%%%%%%%%%%%%%%%%%%%%%%%%%%%%%%%%%%%%%%%%%%%

\clearpage

\appendix
\addcontentsline{toc}{section}{Appendix}
\section*{Appendix}\label{sec:proofs}

\section{Motivating Experiment for all Baselines}
We show the motivating experiment in introduction to pose a problem that such population and separate models have own their limitations when dealing with heterogeneous dataset such as EHR. To recap the results shown in the introduction, we can see that teacher forced RNN misleads predictions on unseen input range (after dashed line), implying the \emph{lack of personalization} and a separate GP model (Personalized GP) on the other hand dismisses the global trend useful for predictions for completely new test patient. This naturally leads us to develop hybrid model that benefits from both complementary aspects of population and separate models, which is in turn to be our model, DME-GP.

\begin{figure}[t]
	\centering
	\includegraphics[width=0.8\linewidth]{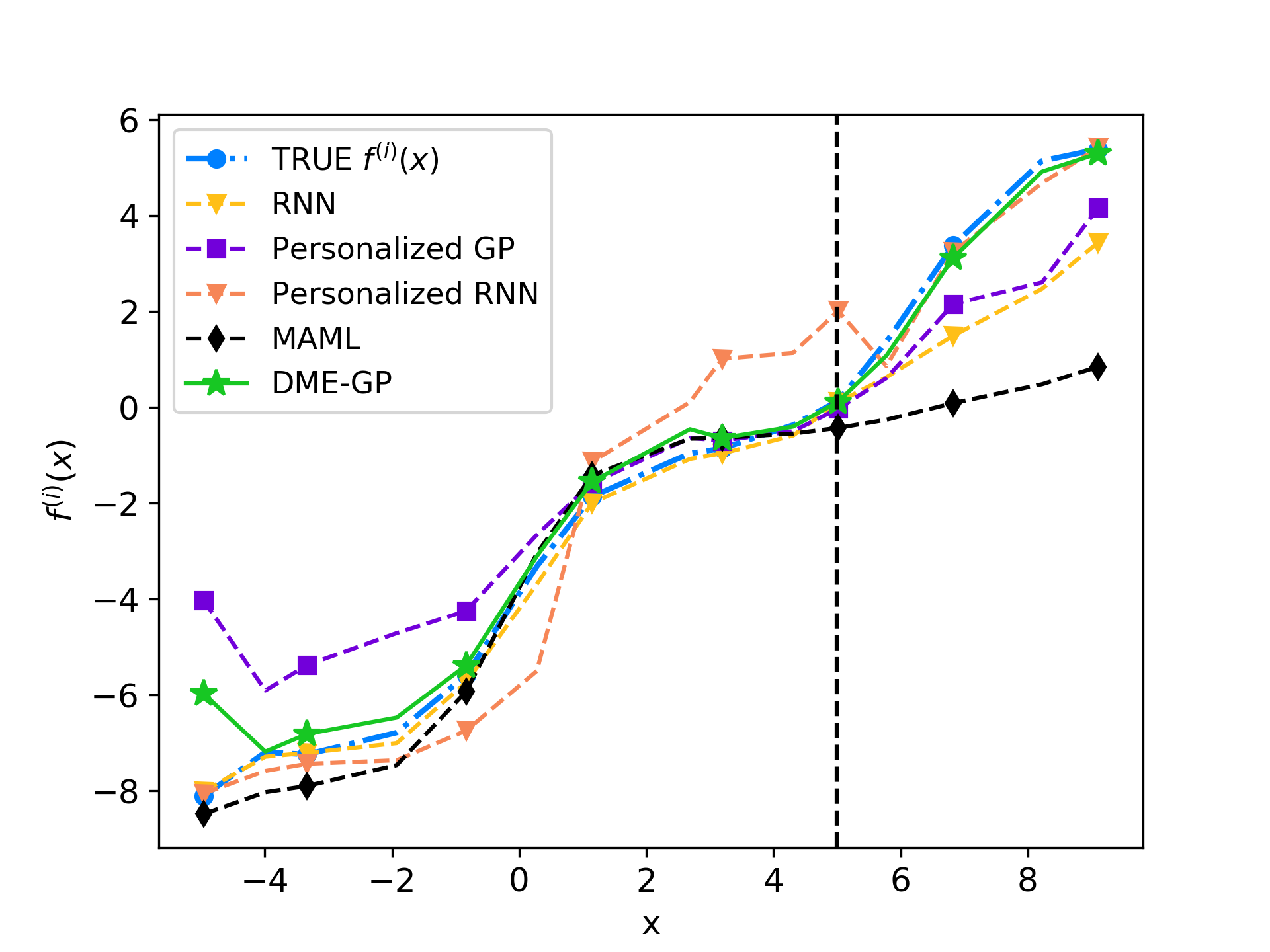}
	\caption{\small \textbf{Motivating experiment.} Results of all baseline models.}
	\label{mot-app}
\end{figure}

We show in appendix the overall results of baseline models for the motivating experiment. In Figure~\ref{mot-app}, we demonstrate the superiority of DME-GP against various baselines. The personalized models (Personalized GP or Personalized RNN), solely constructed for each patient data without any knowledge sharing from training patients, suffer from grasping the global trend in individual function, especially for earlier time points. Specifically, Personalized GP tends to revert to its mean function (zero) and Personalized RNN misleads predictions for earlier stage because of lack of training data (though the model works well later, but this is not a common case in EHR analysis). 
MAML~\citep{maml,meta-sgd}, a typical few shot learning method, successfully benefits from training patients via meta-learning for trained input range (before dashed line), but fails to generalize on unseen input range (after dashed line).

\section{Longer Version of Related Works}

\paragraph{Multiple Gaussian Processes for EHR}
Gaussian Process models have been actively used in the medical applications thanks to its reliability and versatility. However, using the separate formulation of multiple GPs is preferred due to its computational cost. \citet{pgp1} proposed a multiple GPs formulation to handle missing values caused by sensor artifact or data incompleteness, which is common situation in wearable devices. \citet{pgp2} proposed to use a similar model for diagnosis of Alzheimer's disease, where a population-level GP is adapted to a new patient using domain GPs individually. 

\paragraph{Multi-task Gaussian Process for EHR.}
The previous line of work can be understood as multi-task learning in the sense that the parameters of GPs across patients are shared. However, more systematic way of considering multi-task learning with GPs is to directly learn a shared covariance function over tasks. \citet{mtgp4,mtgp2} proposed Multi-task Gaussian Process (MTGP) that constructs large covariance matrix as a Kronecker product of input and task-specific covariance matrices for multi-task learning.
A practical example of applying MTGP in medical situation is given in \citet{mtgp1}.
\citet{mtgp3} proposed another approach that shares covariance matrix structured as the linear model of coregionalization (LMC) framework for personalized GPs, which is generalization of \citet{mtgp5}. \citet{sepsis,sepsis2} made use of MTGP for preprocessing of input data fed into RNN. All of this line of works are based on the multi-task GPs that share huge covariance matrix which makes exact inference intractable. There have been some attempts to utilize mean of GP similar to our approach, proposed by \citet{shared-gp1}, \citet{shared-gp2}, and \citet{mean-nn}. However, our model is constructed in distinctive way where we use flexible deep models for shared mean function to capture complex structures, and more importantly, we explicitly construct a single GP for each patient to reflect individual signal.

\paragraph{Deep learning models with EHR.}
Recurrent neural networks (RNN) have recently been gained popularity as means of learning a prediction model on time-series clinical data such as EHR. \citet{rnn1} and \citet{doctorai} proposed to use RNN with Long-Short Term Memory (LSTM) \citep{lstm} and Gated Recurrent Units (GRU) \citep{gru} respectively for multi-label classification of diagnosis codes given multivariate features from EHR. Moreover, the pattern of missingness, which is typical property of EHR, has been exploited in \citet{rnn3} and \citet{rnn2} by introducing missing indicator and the concept of decaying informativeness. \citet{retain} proposed to use RNN for generating attention on which feature and hospital visit the model should attend to, for building an interpretable model, and demonstrated it on heart failure prediction task. While RNN models have shown impressive performance on real-world clinical datasets, deploying them to safety-critical clinical tasks should be done with caution as they lack the notion of confidence, or uncertainty of prediction.

\paragraph{Combining GPs and deep models.}
Since purely deterministic neural network does not give a confidence about its  prediction, there has been a growing interest in deriving prediction uncertainty using techniques in Bayesian statistics such as \citet{Gal2016Dropout}. One of the direct efforts to accomplish this goal in healthcare is to combine deep architectures and Gaussian Process to benefit from the strengths of both models. \citet{dkgp1} placed GP at the output of a deep network for obtaining more expressive power and similar idea of tweaking covariance function of GP has been studied in \citet{dkgp2,dkgp3}. To achieve the same goal of obtaining a more expressive model, \citet{dgp1,dgp2} formulated a deep network as a stacked Gaussian Processes. Our work also aims to combine GP with a deep neural network, but our composite model is a more effective way to utilize the strengths of the two complementary models. Specifically, we leverage a deep network for capturing the global complex structure in the data, and use GP to capture local variability in the individual instance.

\begin{algorithm}[t]
   \caption{Learning in DME-GP}
   \label{alg:opt}
\begin{algorithmic}
   \STATE {\bfseries Input: $\mathcal{D} = \{ (\X_i, \y_i) \}_{i=1}^{P}, \w, \v, \{ \t_i \}_{i=1}^P$} 
   \WHILE{not converged}
   \STATE Sample minibatch set of patients $\mathcal{B} \subset \mathcal{D}$
   \FOR{each $(\X_j, \y_j) \in \mathcal{B}$}
   \STATE Compute the gradient $\frac{\partial \mathcal{L}_j}{\partial \w}$ in \eqref{Eqn:DERI1_2}
   \STATE Compute the gradient $\frac{\partial \mathcal{L}_j}{\partial \v}$ in \eqref{Eqn:DERI1_2}
   \ENDFOR
   \STATE Update $\w$ and $\v$ based on computed gradients \\ $\{ \frac{\partial \mathcal{L}_j}{\partial \w} \}_{j=1}^{|\mathcal{B}|}$ and $\{ \frac{\partial \mathcal{L}_j}{\partial \v} \}_{j=1}^{|\mathcal{B}|}$
   \FOR{each $(\X_j, \y_j) \in \mathcal{B}$}
   \STATE Compute the gradient $\frac{\partial \mathcal{L}_j}{\partial \t_j}$ in \eqref{Eqn:DERI1_2}
   \STATE Update $\t_j$ based on computed gradients $\frac{\partial \mathcal{L}_j}{\partial \t_j}$
   \ENDFOR
   \ENDWHILE
   \STATE {\bfseries Output: $\w^*, \v^*, \{ \t_i^* \}_{i=1}^P$} 
\end{algorithmic}
\end{algorithm}

\begin{algorithm}[t]
   \caption{Inference for a new patient $j$ in DME-GP}
   \label{alg:inf}
\begin{algorithmic}
   \STATE {\bfseries Input: $(\x_j^*, y_j^*), (\X_j, \y_j), \w^*, \v^*, \t^*$}
   \STATE Update $\t_j$ with fixed number of steps based on the gradients $\frac{\partial \mathcal{L}_j}{\partial \t_j}$ in \eqref{Eqn:DERI1_2}
   \STATE Predict the target $\bar{y}_j$ and uncertainty $\sigma_j^2$ where $p(y_j^* | \x_j^*, \X_j, \y_j)=\mathcal{N}(y_j^* | \bar{y}_j, \sigma_j^2)$ in \eqref{Eqn:GPinference}
   \STATE {\bfseries Output: $\bar{y}_j, \sigma_j^2$} 
\end{algorithmic}
\end{algorithm}

\paragraph{Modeling EHR via Meta-Learning.}
EHR analysis might be casted as few-shot learning problem where each task corresponds to predict time-series output values of each patient given very limited number of personal historical data. As one of the state of the arts to tackle few-shot problem, meta-learning or learning to learn has been widely studied recently. Matching network~\citep{matchingnet} is the one of the pioneer works which introduced episodic-wise training scheme for few-shot classification, based on metric-learning. Many other related works that are based on metric-learning have been developed thereafter~\citep{protonet,relnet}. Another research direction for few-shot problem is based on to learn an optimizer as a meta-learner for a new task, such as Model Agnostic Meta Learning (MAML)~\citep{maml,meta-sgd} and Meta-LSTM~\citep{meta-lstm}, to name a few. However, it is not trivial to apply episodic training in case of EHR analysis due to its times-series property (even in our experiments a naive casting to meta-learning without episodic training performs poorly in most cases), and EHR analysis within meta-learning framework has not been studied thoroughly so far. 

\begin{table}[t]
\caption{\small \textbf{Data Statistics.} The number of patients ($P$) and total samples for each disease. Maximum, minimum, and average time steps for the patient are 12, 3, and 6.7 respectively.}
\label{DATADESC}
\centering
\begin{tabular}{lcc}
\toprule
Diseases & $P$ & Total Samples \\
\midrule
Alcoholic Fatty Liver		& 1,593 & 11,050\\
Atherosclerosis				& 6,953 & 46,206\\
Emphysema    				& 1,513 & 9,978\\
Liver Cirrhosis   			& 2,246 & 15,228\\
Alcoholic Hepatitis   		& 1,118 & 7,537\\
Arrhythmia     				& 364 & 2,432\\
Fatty Liver    				& 9,273 & 62,938\\
Heart Failure  		 		& 5,018 & 32,465\\
Hepatic Failure 			& 446 & 2,968\\
Hepatitis B 				& 450 & 3,001\\
Myocardial Infarction 		& 1,926 & 12,967\\
Toxic Liver Disease 		& 2,027 & 13,638\\
\bottomrule
\end{tabular}
\end{table}

\section{Global Component as a Mixture of Experts}\label{subsec:MIX}
A mixture of experts model \cite{moe1} is built upon the assumption that complex problems may contain many sub-problems which can be efficiently resolved by assigning each sub-problem to certain expert which is good at solving it. %Natural extension of this concept is aligned to our global mean function. 

We hypothesize that the global mean function $\mu(\cdot)$ can be divided into several experts since the patients can be clustered into similar groups. For example, the patients can be clustered into the groups having different range of ages or different range of where they live. Hence, we posit %This concept can be expressed by the concept of a mixture of experts model as explained above, which is,
\begin{align}
	\mu(\x_t) = \sum_{j} g(j|\x_t) \mu_j(\x_t)
\end{align}
where $g(\cdot)$ is a gate function for which the sum of probability to choose a certain expert is one, and $\mu_j(\cdot)$ is an expert in charge of some portion of input space.

\section{Algorithms for Training and Inference in DME-GP}\label{sec:algorithm}
We present algorithms for training and inference in DME-GP as shown in Algorithm~\ref{alg:opt} and Algorithm~\ref{alg:inf}. Note that for inference of a new patient, we find that it is beneficial to have shared covariance parameters in learning as the initial states of the covariance parameters for the new patient in adaptation.

\section{Data Statistics}\label{sec:DATADESC}
Data statistics on our private datasets from EHR (from National Health Insurance Service; NHIS)) are given in Table~\ref{DATADESC}.

\begin{table}[t]
\caption{\small \textbf{Mixture of Experts Study.} Performance (in terms of Area Under the ROC Curve; AUC) comparisons among mixture models. MIX1, 2, 4 have experts 1, 2, 4 numbers respectively.}
\label{MIXTURE}
\centering
\begin{tabular}{lccc}
\toprule
%\multirow{2}{*}{Diseases} & \multicolumn{3}{c}{Mixture models} \\
Diseases & mix1 & mix2 & mix4 \\
\midrule
Alcoholic Fatty Liver	& $\bm{0.801}$	& $\bm{0.801}$	& 0.798 \\
Atherosclerosis			& 0.74			& $\bm{0.744}$	& 0.735 \\
Emphysema    			& 0.742			& 0.782			& $\bm{0.799}$ \\
Liver Cirrhosis   		& $\bm{0.922}$	& 0.920			& 0.919 \\
Alcoholic Hepatitis   	& 0.852			& 0.838			& $\bm{0.868}$ \\
Arrhythmia     			& 0.74 			& 0.723			& $\bm{0.823}$ \\
Fatty Liver    			& 0.731			& $\bm{0.732}$	& $\bm{0.732}$ \\
Heart Failure  	 		& 0.759			& $\bm{0.78}$ 	& 0.766 \\
Hepatic Failure 		& $\bm{0.738}$	& 0.734			& 0.619 \\
Hepatitis B 			& 0.489 		& 0.464			& $\bm{0.531}$ \\
Myocardial Infarction 	& 0.826 		& $\bm{0.854}$	& 0.81 \\
Toxic Liver Disease 	& $\bm{0.698}$ 	& 0.697			& 0.678 \\
\bottomrule
\end{tabular}
\end{table}

\section{Training Details in Experiments}
We train all models on 70\% of the full dataset separated by patients, validate the models on 10\% for searching appropriate hyper-parameters, and use the remaining 20\% for evaluation. We apply two regularization methods to avoid overfitting of deep models: early stopping and $\ell_2$-penalization. We also use the dropout technique for deep models to improve performances \citep{dropout}. We optimize the parameters of the models by stochastic gradient descent with ADAM optimizer \citep{adam}. For the models composed of GPs, we use a squared exponential (or RBF) kernel function with automatic relevance determination (ARD) to fully utilize the property of the universal approximator \citep{kernel1}. For application of these models to binary classification tasks, we use a variational method \citep{vgp,svgp} to handle intractability of inference caused by non-Gaussian likelihood.

\section{Additional Experiments on Disease Risk Prediction Tasks}

\subsection{Global Component as a Mixture of Experts}
We also evaluate mixture models having various number of experts as a global mean function. In particular, we consider the cases of being 1, 2, 4 number of experts. We use MLP with one hidden layer as an expert (here a part of a global mean function) and softmax classifier as a gate function.
As mentioned in Section~\ref{subsec:MIX}, instead of having a single mean function we can use a mixture of experts to model the mean function, as patients could be divided into several precision cohorts. We present the experiments on this mixture of experts in Table~\ref{MIXTURE}. The results show that this mixture of experts can be effective for certain types of diseases, such as Arrhythmia and Hepatitis B, while further investigation is required to see how the cohorts are formed.

\subsection{Reliability Study}\label{sec:cali}
We report the results of reliability study for all disease risk prediction tasks. Again, we define a \emph{hard patient} to denote a patient who has a positive label \emph{only} at the prediction time $T_i$ but never has positive labels in his/her historical data. The results in Table~\ref{cali} compare the performances of the models when we exclude positive samples belonging to \emph{hard patients}. On the other hand, Table~\ref{cali2} shows comparison of the models when we only include a set of \emph{hard patients} as positive samples. AUC scores in these experiment are obtained with the model with best-fit hyper-parameters. 

\begin{table}[t]
\caption{\small \textbf{Reliability Study.} Performance (in terms of Area Under the ROC Curve; AUC) comparisons when we exclude a set of \emph{hard patient} who has a positive label at $T_i$ but negative labels for others.}
\label{cali}
\centering
\begin{tabular}{lccc}
\toprule
Diseases & DME-GP & RNN & RETAIN \\
\midrule
A. Fatty Liver			& $\bm{0.998}$	& 0.856	& 0.861 \\
Atherosclerosis			& $\bm{1}$	& 0.683	& 0.768 \\
Emphysema    			& $\bm{0.991}$	& 0.785	& 0.765 \\
Liver Cirrhosis   		& $\bm{0.989}$	& 0.928	& 0.922 \\
Alcoholic Hepatitis   		& $\bm{0.981}$	& 0.881	& 0.866 \\
Arrhythmia     			& $\bm{0.992}$	& 0.639	& 0.656 \\
Fatty Liver    			& $\bm{0.996}$	& 0.741	& 0.742 \\
Heart Failure  	 		& $\bm{0.983}$	& 0.853	& 0.849 \\
Hepatic Failure 			& $\bm{1}$ 	& 0.584	& 0.634 \\
Hepatitis B 			& $\bm{1}$ 	& 0.695	& 0.491 \\
M. Infarction 			& $\bm{0.982}$	& 0.943	& 0.944 \\
Toxic Liver Disease 		& $\bm{1}$ 	& 0.593	& 0.597 \\
\bottomrule
\end{tabular}
\end{table}

\begin{table}[t]
\caption{\small \textbf{Reliability Study.} Performance (in terms of Area Under the ROC Curve; AUC) comparisons when we only include a set of \emph{hard patient} for positive samples.}
\label{cali2}
\centering
\begin{tabular}{lccc}
\toprule
Diseases & DME-GP & RNN & RETAIN \\
\midrule
A. Fatty Liver			& 0.755	& $\bm{0.78}$	& 0.767 \\% & 15.9\% \\
Atherosclerosis			& $\bm{0.691}$	& 0.653	& 0.676 \\% & 30.2\% \\
Emphysema    			& $\bm{0.783}$	& 0.754	& 0.734 \\% & 26.2\% \\
Liver Cirrhosis   		& $\bm{0.837}$	& 0.828	& 0.817 \\% & 6.6\% \\
Alcoholic Hepatitis   	& 0.798	& $\bm{0.809}$	& 0.75 \\% & 11.4\% \\
Arrhythmia     			& 0.591	& 0.453	& $\bm{0.678}$ \\% & 51.2\% \\
Fatty Liver    			& 0.653	& $\bm{0.674}$	& 0.665 \\% & 34.2\% \\
Heart Failure  	 		& 0.734	& $\bm{0.768}$	& 0.766 \\% & 15.2\% \\
Hepatic Failure 		& $\bm{0.666}$ & 0.621	& 0.617 \\% & 57.7\% \\
Hepatitis B 			& 0.547 & 0.62	& $\bm{0.722}$ \\% & 43.9\% \\
M. Infarction 			& $\bm{0.79}$	& 0.783	& 0.781 \\% & 4.0\% \\
Toxic Liver Disease 	& 0.618 & 0.635	& $\bm{0.643}$ \\% & 67.5\% \\
\bottomrule
\end{tabular}
\end{table}

\subsection{Reliability Study via ROC Curve Analysis}
\begin{figure*}[t]
	\centering
	\includegraphics[width=1.0\linewidth]{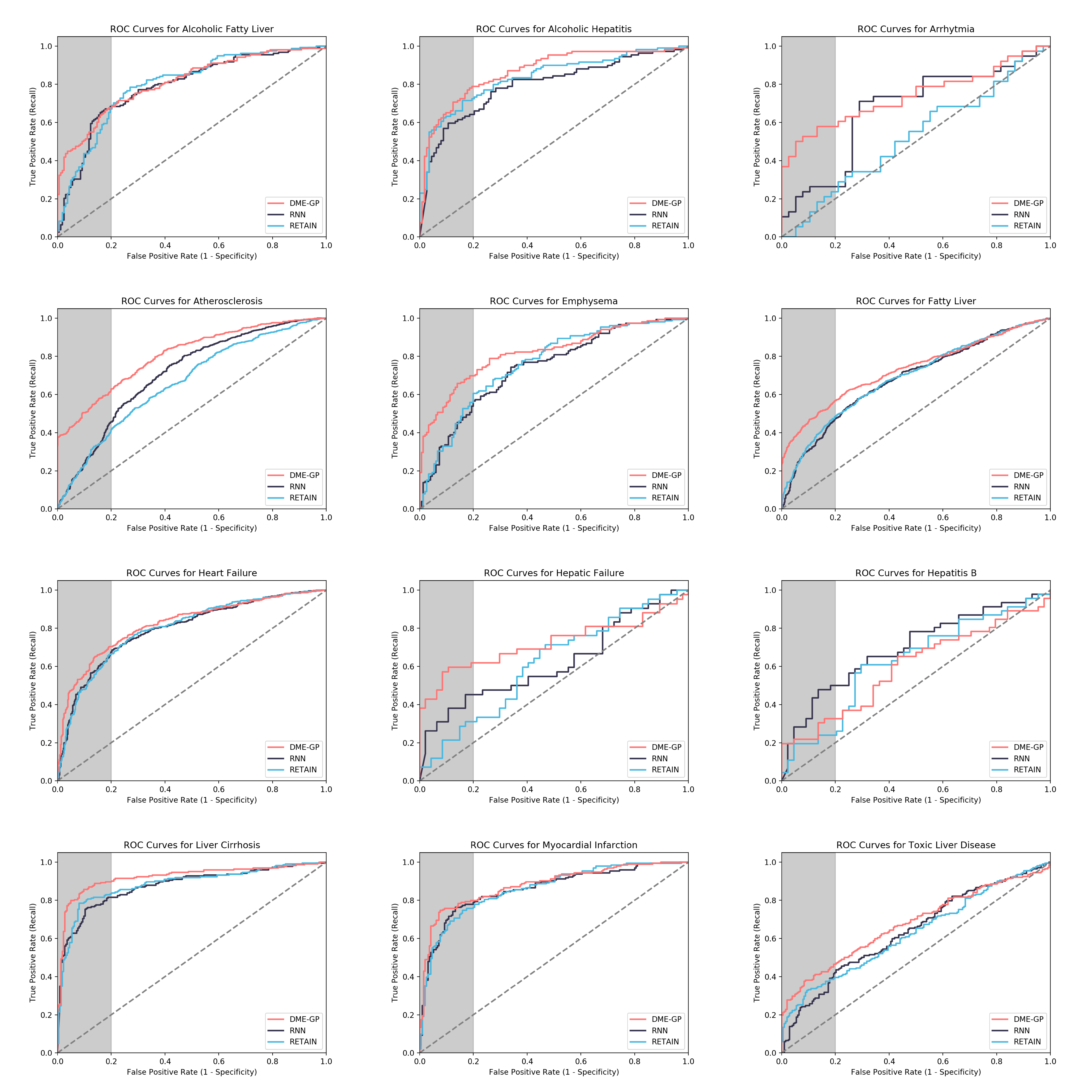}
	\caption{\textbf{Reliability Study.} ROC curves of DME-GP and baselines in risk prediction experiment. We present curves for all diseases and the name of each of which is shown in the title of subplot. Grey-shaded area represents early retrieval stage.}
	\label{roc_fig_app}
\end{figure*}

We further study the patterns of ROC curves for all diseases to measure the predictive reliability of our model and compare to other baseline models. Specifically in ROC curves, we focus on early stage of x-axis (1-specificity) shaded as grey area in Figure~\ref{roc_fig_app}, where it measures \emph{recall} of highly probable positive cases, called early retrieval \citep{roc1,roc2}. In typical medical cases including our risk prediction tasks, \emph{recall} really matters since detection of positive cases are important, especially for intensive ones in early retrieval area. For most of diseases, DME-GP forms recall curves above the ones of baselines in early stage, which means the model can reliably detect urgent patients in ranked way. This ROC curves analysis shows consistent results with the results shown in the previous reliability study and has strong connection with it. This reliability of our model will also allow proper involvement of human medical staff.

\end{document}